\documentclass[preprint,12pt]{elsarticle}

\usepackage{dsfont,amssymb,amsmath,graphics,graphicx,fancyhdr,float,verbatim,
ifthen,ifpdf,algorithmic,multirow,array,color, enumerate,theorem}
\usepackage[english]{babel}
\usepackage[perpage]{footmisc}
\usepackage[ruled]{algorithm}
\usepackage[subnum]{cases}

\usepackage{natbib}
\biboptions{authoryear, round}

\graphicspath{{EpsFigs/}}

\newcommand{\bB}{\mathbf{B}}
\newcommand{\bO}{\mathbf{0}}

\newcommand{\be}{\mathbf{e}}

\newcommand{\bt}{\mathbf{t}}

\newcommand{\bw}{\mathbf{w}}
\newcommand{\bW}{\mathbf{W}}

\newcommand{\bX}{\mathbf{X}}

\newcommand{\bY}{\mathbf{Y}}
\newcommand{\bz}{\mathbf{z}}

\newcommand{\bst}{\boldsymbol{t}}

\newcommand{\bsv}{\boldsymbol{v}}
\newcommand{\bsw}{\boldsymbol{w}}

\newcommand{\bsy}{\boldsymbol{y}}

\newcommand{\cL}{\mathcal{L}}

\newcommand{\cO}{\mathcal{O}}

\newcommand{\bsbeta}{\boldsymbol{\beta}}

\newcommand{\bsSigma}{\boldsymbol{\Sigma}}

\newcommand{\bstheta}{\boldsymbol{\theta}}

\newcommand{\E}{\mathbb{E}}

\newcommand{\R}{\mathbb{R}}
	
\newcommand{\N}{\mathcal{N}}

\journal{Neurocomputing Elsevier}

\begin{document}

\begin{frontmatter}

\title{Joint segmentation of multivariate time series with hidden process regression for human activity recognition}

\author[LSIS-USTV,LSIS-AMU]{\hspace*{-0.15cm} F. Chamroukhi \corref{cor1}} \ead{faicel.chamroukhi@univ-tln.fr}
\cortext[cor1]{Corresponding author: Faicel Chamroukhi\\ Universit\'e de Toulon, LSIS, UMR CNRS 7296 \\  B\^atiment R, BP 20132 - 83957 La Garde Cedex, France \\Tel: +33(0) 4 94 14 20 06\\Fax: +33(0) 4 94 14 28 97 }

\author[Upec]{S. Mohammed} \ead{samer.mohammed@u-pec.fr}
\author[Upec]{D. Trabelsi} \ead{dorra.trabelsi@u-pec.fr}
\author[Ifsttar]{L. Oukhellou} \ead{latifa.oukhellou@ifsttar.fr}
\author[Upec]{Y. Amirat} \ead{amirat@u-pec.fr}

\address[LSIS-USTV]{Universit\'e de Toulon, CNRS, LSIS, UMR 7296, 83957 La Garde, France} 
\address[LSIS-AMU]{Aix Marseille Universit\'e, CNRS, ENSAM, LSIS, UMR 7296, 13397 Marseille, France}
\address[Upec]{University Paris-Est Cr\'eteil (UPEC), LISSI, 122 rue Paul Armangot, 94400, Vitry-Sur-Seine, France}
\address[Ifsttar]{University Paris-Est, IFSTTAR, GRETTIA, F-93166 Noisy-le-Grand, France}

\begin{abstract}
The problem of human activity recognition is central for understanding and predicting the human behavior, in particular in a prospective of assistive services to humans, such as health monitoring, well being, security, etc. There is therefore a growing  need to build accurate models which can take into account the variability of the human activities over time (dynamic models) rather than static ones which can have some limitations in such a dynamic context. In this paper, the problem of activity recognition is analyzed through the segmentation of the multidimensional time series of the acceleration data measured in the 3-d space using body-worn accelerometers. The proposed model for automatic temporal segmentation is a specific statistical latent process model which assumes that the observed acceleration sequence is governed by sequence of hidden (unobserved) activities. More specifically, the proposed approach is based on a specific multiple regression model incorporating a hidden discrete logistic process which governs the switching from one activity to another over time. The model is learned in an unsupervised context by maximizing the observed-data log-likelihood via a dedicated expectation-maximization (EM) algorithm. We applied it on a real-world automatic human activity recognition problem and its performance was  assessed by performing comparisons with alternative approaches, including  well-known supervised static classifiers and the standard hidden Markov model (HMM). The obtained results are very encouraging and show that the proposed approach is quite competitive even it  works in an entirely unsupervised way and does not requires a feature extraction preprocessing step.

\end{abstract}

\begin{keyword} 
Human activity recognition \sep inertial sensors \sep acceleration data  \sep times series segmentation \sep hidden process regression  \sep unsupervised learning \sep expectation-maximization algorithm 
\end{keyword}

\end{frontmatter}

\section{Introduction}
\label{sec::Introduction}
The problem of human activity recognition is central for understanding and predicting the human behavior, in particular for a prospective of assistive services to elderly people. In fact, assistive services related to the  aging population has gained an increasing attention in the last decades due the fact that this population is increasing and having more socio-economic impact.  
The aim is therefore to facilitate the daily lives of elderly or dependent people at home, increase their autonomy and improve their safety. The emergence of novel adapted technologies such as wearable and ubiquitous technologies is becoming a privileged solution to provide assistive services to humans, such as health monitoring, well being, security, etc. For example, the recognition of the activity from  data measured by sensors can play important role in minimizing the risk of human fall \citep{Hirata, Kangas2008, Noury, Lindemann2005}. In addition, remote monitoring based systems can reduce considerably the amount of admission to hospitals by early detection of gradual deterioration in elderly health status \citep{Scanaill2006}.

In the context of activity recognition from measurements, the main measurement techniques used to quantify human activities are based on the use of inertial sensors \citep{Altun,Trabelsi_wIROS, Preece, Kavanagh,JuhaEtall06}. Other sensors are also used to recognize human activities such as video-based systems \citep{CappozzoA, CappozzoB, Aminian1999}, goniometers \citep{Kostov1995}, electromyography (EMG) \citep{hussein2002}, etc. However, the video-based approach is less adapted regarding particularly the personal privacy aspects.  In addition, technological advances and miniaturization of wearable inertial sensors made easier the collection of data for a wide range of human activities. As a consequence, the technique based on wearable-sensors  has gained more attention for activity recognition as well as in many application domains including
 medical applications such as rehabilitation program and medical diagnosis \citep{Jovanov05}. 
In the context of werable-sensors based activity recognition, the accelerometers are the most commonly used inertial sensors due to the advances in micro-electromechanical systems technology which have greatly promoted the use of accelerometers thanks to the considerable reduction in size (miniaturazation), in cost and in energy consumption. These sensors have shown satisfactory results to measure the human activities in both laboratory, clinical and free-living environment settings \citep{Mathie}. More specifically, the tri-axial accelerometers are more privileged \citep{Yang} with respect to uniaxial accelerometers.  

The studied activities can be either static ones (postures) such as standing, sitting, lying or dynamic ones (movements) such as standing up, sitting down, walking, running, climbing stairs, etc. 
In general, the activity recognition from acceleration data is preceded by a preprocessing step of feature extraction.  For example features, such as the mean, the standard deviation, the skewness, the kurtosis, etc. can be extracted from original data and used afterwards as classifier inputs \citep{Altun, Ravi, Yang}. However, these methods are static ones as the models do not exploit the temporal dependence of the data. An alternative approach may use features such as zero-velocity crossings (ZVC) \citep{Ajo_Mataric_Jenkins}. 
A method for dynamic signal segmentation for activity recognition has been proposed in \citep{Kozina}; However, it has lower accuracy with transition instances in learning and testing set. Additional details on classification approaches for human activity recognition can be found in the recent reviews \citep{Yang,Altun,Preece,Kavanagh}. Among the techniques used in the literature for the human activity classification, one can distinguish those using supervised machine learning-based approaches. These techniques provide a statistical association of a given activity feature to a possible class. For example, one can cite $K$-nearest neighbor ($K$-NN) classification algorithm \citep{Foerster},  the naive Bayes classifier \citep{Long} and the support vector machines (SVM) \citep{Lau}. For the unsupervised context, one can cite the approach based on Gaussian mixture models (GMM) \citep{Allen} and the one based on hidden Markov model (HMM) \citep{Jonathan-W-IROS11} or HMM with GMM emission probabilities \citep{Mannini}.

\begin{figure}[!h]
\centering
\includegraphics[height=6cm, width=9cm]{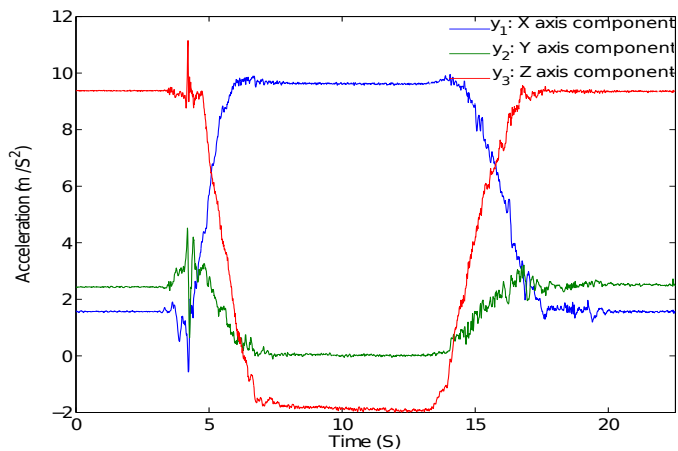}
\caption{\label{fig : Acc data for illustration}Example of acceleration times series.}
\end{figure}

In this study, the human activities are detected over time through the segmentation of the accelerations time series. The data are measured over time during the activity of a given person, and at each time step,  the three acceleration components, which we denote by $(y_1,y_2,y_3)$, are recorded. These data consist therefore in multidimensional time series with several regime changes over time, each regime is associated with an activity (e.g., see Figure \ref{fig : Acc data for illustration}). The problem of activity recognition can therefore be reformulated as the one of a joint segmentation of multidimensional time series, each segment is associated with an activity.
 
The proposed statistical approach is dedicated to temporal segmentation by including a hidden process where the process probabilities change over time according to the most likely activity. The approach performs in an unsupervised context from raw acceleration data. More specifically, the proposed model consists in a multiple regression model governed by a discrete hidden process. The configuration of the hidden process, at each time step, corresponds to an activity described by a regression model. The hidden process configuration depends on time and the regression model parameters are time-varying according to the most likely posture. Furthermore, although only the acceleration measurements are available, the classes of activities being unknown, the unknown class labels are stated in the model as latent variables. In this way, we formulate an efficient and computationally tractable statistical latent data model. The resulting model is therefore a kind of latent data model which is particularly well adapted for performing unsupervised activity recognition.  
Let us recall that, from a statistical prospective, latent data models \citep{spearman1904} aim at representing the distribution $p(\bsy)$ of the observed data in terms of a number of latent variables $z$. In this context of activity recognition, the observed data are the acceleration data and the latent data represent the activities (the class labels). Mixture models \cite{mclachlanFiniteMixtureModels} and hidden Markov models \citep{rabiner} are two well-known widely used examples of such models. The unsupervised  learning task for the proposed approach is achieved by maximizing the observed-data log-likelihood via a dedicated iterative algorithm known as the expectation-maximization (EM) algorithm. 

\section{Related work on activity recognition and time series segmentation} 
\label{sec: related work}
In this section, we briefly describe some classification approaches which are commonly used for the problem of human activity recognition. These approaches represent statistical, neural and distance-based techniques.
First one can cite the naive Bayes classifier (e.g. see \cite{mitchell_book_1997}) which is a  supervised probabilistic classifier based on Bayes' theorem. To simplify the model estimation procedure, this classifier assumes the naive assumption that the acceleration measurements are conditionally independent given the activities.  Neural approaches have also been used to perform activity recognition \citep{Yang08}, in particular the multilayer perceptron (MLP), proposed by \citep{Rumelhart} are widely used approaches for supervied classification. The model training is performed by minimizing a cost function between the estimated and the desired network outputs (in this case the activity labels) using a gradient descent.  
Support vector classifiers (SVC) introduced by Vapnik (e.g., see \cite{vapnik99}) are derived from statistical learning theory and are proved to be very efficient supervised classifiers. They minimize an empirical risk and at the same time attempt at finding a separating hyperplane that maximizes the margin to the classes. %
The $k$-Nearest Neighbor ($k$-NN) \citep{CoverEtHart} is a non parametric distance-based supervised classifier which is widely used because of its efficiency and simplicity of implementation. It classifies a data example according to a majority vote of its $k$ nearest data examples in the sense of a chosen metric, typically an Euclidean distance.
A recent Dynamic Time Warping (DTW) approach with semantic attributes (SDTW) has been proposed in  \cite{Pogorelc2012} and increases the classification accuracy compared to some classical machine learning techniques for activity classification. 

%
The classification approaches described above are supervised and require therefore a labeled collection of data to be trained. Besides, they do not exploit the temporal dependence in their model formulation as they are static approaches. The hidden Markov model (HMM) 
 is a latent data dynamic model which is well adapted for modeling sequences and has shown its performance especially in speech recognition. It assumes that the observed sequence is governed by a hidden state (activity) sequence where the current activity depends on the previous one rather than an independence hypothesis as in the previously described classifiers. The HMM can also be used for temporal activity recognition \citep{Mannini,Jonathan-W-IROS11}.

In this paper, we formulate the problem of activity recognition as the one of a
joint segmentation of multidimensional time series. 
The general problem of time series segmentation has taken great interest from different communities, including statistics, detection, signal processing, machine learning, and robotics namely activity recognition, etc. Earlier contributions in the subject were taken from a statistical point of view 
\citep{McGee,Brailovsky,  rabiner, Keogh93segmentingtime, Nikiforov, Fridman, Kohlmorgen02adynamic,  DobigeonAndAll2007, KulicAndAll2008, LinAndKulic2011}. 
In \citep{McGee,Brailovsky}, the time series segments are modeled as several regime changes over time, each regime is assumed to be a noisy constant function and corresponds to a segment.  
This leads to the standard piecewise regression model proposed by \citep{McGee,Brailovsky}. In the piecewise regression model extended in \citep{Brailovsky}, the data are partitioned into several segments, each segment being characterized by its mean polynomial curve and its variance. However, the parameter estimation in such a  method requires the use of dynamic programming algorithm \citep{Bellman,Stone} which may be computationally expensive especially for time series with a large number of observations. Moreover, the standard piecewise regression model usually assumes that noise variance is uniform in all the segments (homoskedastic model).
In \citep{Nikiforov}, the problem is stated as a detection problem via hypothesis testing. This type of approach in general requires a detection threshold to reject the null hypothesis. In addition, the hypothesis testing is often used in a binary setting, the problem of multiple hypothesis testing, which is the case in this multi-class segmentation problem, is not very common while one can take each two hypotheses independently to be tested. 
Alternative approaches come from the machine learning community, one can cite for example the hidden Markov model \citep{rabiner, LinAndKulic2011}  which is a well-known approach that assumes that the data are arranged in an observation sequence generated by a hidden state sequence. For an activity recognition problem, each state represents an activity. The HMM can be trained in a batch-mode with a fixed number of states as well as in an online mode with varying number of states as in \citep{Kohlmorgen02adynamic}. However, in this paper, the number of activities is assumed to be fixed and the model acts in a batch mode.
 Another way for time series modeling is to use a hidden Markov model  regression  \citep{Fridman} which is a formulation of the standard HMM into an univariate regression context. This can also be extended to a multidimensional regression setting as in \citep{Trabelsi_wIROS} which was also applied on time series acceleration data. Another  statistical Bayesian approach that uses Gibbs sampling was also proposed for a 
joint segmentation of multidimensional astronomical time Series \citep{DobigeonAndAll2007}. While this approach is non-parametric, it requires Gibbs sampling to evaluate the posterior distribution of the model and to estimate the hyperparameters of the model. This can be very limiting in terms of computational time. For the proposed approach, the posterior distribution is computed analytically and does not require sampling. In addition, while the proposed approach is parametric as in its current formulation the number of activities is given, as we shall specify it, one can use a model selection procedure to select the optimal number of activities based on some information criteria.
 
For the particular problem of human activity recognition, the statistical latent data models, such as standard HMMs \citep{LinAndKulic2011} or dynamic HMMs  \citep{Kohlmorgen02adynamic} have shown their performance in terms of activity recognition based on times series segmentation. Another approach uses a combined segmentation and clustering approach, and is also based on HMMs as in \citep{KulicAndAll2008}. 
These approaches can be used namely in an online mode \citep{Kohlmorgen02adynamic, KulicAndAll2008, KulicAndNakamura2008}. In this paper, we propose an alternative approach to HMMs in a context of a multiple regression in a batch-mode, it can be as well adapted to an online learning framework.
The approach we propose to perform temporal segmentation of multivariate time series is based on an alternative to the Markov process in the HMM regression model \citep{Fridman, Trabelsi_wIROS}. It also directly uses the raw acceleration data rather than performing feature extraction and feature selection as in \citep{Altun, Ravi, Yang}. 
%
%
In the next section, we will present the proposed model for joint segmentation of multiple time series.

\section{The multiple regression model with a hidden logistic process (MRHLP)} 
\label{sec: M-rhlp model}

Let $\bY=(\bsy_1,\ldots,\bsy_n)$ be a time series of $n$  multidimensional observations  $\bsy_i = (y_{i}^{(1)},\ldots,y_{i}^{(d)})^T\in \R^d$  regularly observed at the time points $\bt=(t_1,\ldots,t_n)$. Assume that the observed time series is generated by a $K$-state hidden process and let $\bz=(z_1,\ldots,z_n)$ be the unknown (hidden) state sequence associated with the time series $(\bsy_1,\ldots,\bsy_n)$  with $z_i \in \{1,\ldots,K\}$. Each observation $\bsy_i$ represents an acceleration measurement and the corresponding state $z_i$ represents its associated activity.

The proposed approach extends the regression model with a hidden logistic process (RHLP) \citep{chamroukhi_et_al_NN2009} which is concerned with univariate time series, to the multivariate case. Furthermore, the general model formulation presented in this paper includes an additional point, that is the possibility to train the polynomial dynamics with different orders rather than assuming a common order for all the polynomials.  

In this way, the model offers more flexibility allowing the capture of nonlinearities between different activity acceleration data.
 Let us first recall that the basic univariate RHLP model is stated as follows:
\begin{equation}
y_i = \bsbeta^T_{z_i}\bt_i + \sigma_{z_i}\epsilon_i \quad ; \quad\epsilon_i \sim \N(0,1) , \quad (i=1,\ldots,n)
\label{eq:  rhlp regression model}
\end{equation}
where ${z_i}$ is a hidden discrete-valued variable taking its values in the set $\{1,\ldots,K\}$. The variable $z_i$ controls the switching from one activity to another for $K$ activities at each time $t_i$. The vector $\bsbeta_{z_i}=(\beta_{z_i0},\ldots,\beta_{z_i p})^T$ is the one of regression coefficients of the polynomial regression model $z_i$, $\bt_i=(1, t_i,t_i^{2} \ldots, t_i^{p})^T$ is the $p+1$ dimensional covariate vector at time $t_i$ and the finite integer $p$ represents the polynomial order. The RHLP model assumes that the hidden sequence $\bz=(z_1,\ldots,z_n)$ is a hidden logistic process detailed in \citep{chamroukhi_et_al_NN2009} and which will be recalled  subsequently. 

For the multiple regression case studied here, the model is reformulated as a set of several polynomial regression models (RHLP) for univariate data and is stated as follows:
\begin{eqnarray}
y^{(1)}_i &=& \bsbeta^{(1)T}_{z_i}\bt_i + \sigma^{(1)}_{z_i}\epsilon_i \nonumber \\
y^{(2)}_i &=& \bsbeta^{(2)T}_{z_i}\bt_i + \sigma^{(2)}_{z_i}\epsilon_i \nonumber \\
\vdots & & \vdots \nonumber \\
y^{(d)}_i &=& \bsbeta^{(d)T}_{z_i}\bt_i + \sigma^{(d)}_{z_i}\epsilon_i 
\label{eq:  M-rhlp regression model}
\end{eqnarray}
where $d$ represents the dimension of the time series and the latent process $\bz$ simultaneously governs all the univariate time series components. This allows therefore for a joint segmentation of the acceleration data over time. In (\ref{eq:  M-rhlp regression model}),  we have $\bt_i=(1, t_i,t_i^{2} \ldots, t_i^{p_{k}})^T$ with $p_{k}$ is the degree of the polynomial model associated with the class $z_i = k$.
 The model (\ref{eq:  M-rhlp regression model}) can be rewritten in a matrix form as:
 \begin{equation}
\bsy_i =  \bB_{z_i}^T  \bt_i + \be_i \quad ; \quad\be_i \sim \N(\bO,\bsSigma_{z_i}) , \quad (i=1,\ldots,n)
\label{eq:  M-rhlp matrix form}
\end{equation}
where $\bsy_i=(y^{(1)}_i,\ldots,y^{(d)}_i)^T$ is the $j$th observation in $\R^d$, $\bB_k = \left[\bsbeta^{(1)}_{k},\ldots, \bsbeta^{(d)}_{k} \right]$ is a $(p+1)\times d$ dimensional matrix of the  multiple regression model parameters associated with the regime (class) $z_i=k$ and $\bsSigma_{z_i}$ 
its corresponding covariance matrix.

The probability distribution of the process  $\bz=(z_1,\ldots,z_n)$, that allows for the switching from one regression model to another, and therefore from a posture or a movement to another,  
is assumed to be logistic. Formally, the hidden logistic process assumes that the variables $z_i$ $(i=1,\ldots,n)$, given the vector $\bst=(t_1,\ldots,t_n)$, are generated independently according to the multinomial distribution {\small$\mathcal{M}(1,\pi_{1}(t_i;\bw),\ldots,\pi_{K}(t_i;\bw))$}, where:
\begin{equation} 
\pi_{k}(t_i;\bw)= p(z_i=k|t_i;\bw)=\frac{\exp{(\bsw^T_k \bsv_i})}{\sum_{\ell=1}^K\exp{(\bsw^T_\ell \bsv_i})},
\label{eq: multinomial logit}
\end{equation}
is the logistic transformation of a linear function of the time-dependent covariate vector $\bsv_i=(1,t_i,,\ldots,t^u_i)^T$, $\bsw_{k}=(w_{k0},w_{k1},\ldots,w_{kq})^T$ is the $u+1$-dimensional coefficients vector associated with $\bsv_i$ and $\bw = (\bsw_1,\ldots,\bsw_K)$.  
The RHLP process is well-adapted for capturing both abrupt and/or smooth changes of activities thanks to the flexibility of the logistic distribution. The relevance of the logistic process in terms of flexibility of transition is detailed in \cite{chamroukhi_et_al_NN2009}.
  
Now consider the distribution of the observed data for the proposed model. It can be easily shown that the observation $\bsy_i$ at each time point $t_i$ is distributed according to the following conditional normal mixture density:
\begin{eqnarray}
p(\bsy_i|t_i;\bstheta)&=&\sum_{k=1}^K \pi_{k}(t_i;\bw) \N\big(\bsy_i;\bB^T_k\bt_i,\bsSigma_k\big) ,
\label{melange}
\end{eqnarray}
where $\bstheta=(\bw,\bB_1,\ldots,\bB_K,\bsSigma_1,\ldots,\bsSigma_K)$  is the unknown parameter vector to be estimated. 

\subsection{Parameter estimation by a dedicated EM algorithm}
\label{ssec: parameter estimation for the RHLP}

The parameter $\bstheta$ is estimated using the maximum likelihood method. As in the classic regression models, we assume that, given $\bst =(t_1,\ldots,t_n)$,  the $\epsilon_i$ are independent. This also implies the independence of $\bsy_i$ $(i=1,\ldots,n)$ given the time vector $\bt$. The log-likelihood of $\bstheta$ for the observed data $\bY = (\bsy_1,\ldots,\bsy_n)$ is therefore written as:
\begin{eqnarray}
\cL(\bstheta;\bY,\bt)&=&\sum_{i=1}^{n}\log\sum_{k=1}^K \pi_{k}(t_i;\bw)\N\big(\bsy_i;\bB^T_k\bt_i,\bsSigma_k\big) .
\end{eqnarray}
The maximization of this log-likelihood cannot be performed in a closed form  since it results in a complex nonlinear function due to the logarithm  of the sum. However, in this context of latent data model, the expectation-maximization (EM) algorithm \citep{mclachlanEM,dlr} is particularly adapted for maximizing  the log-likelihood. 
To derive the EM algorithm for the proposed model, we first give the complete-data log-likelihood:  
\begin{eqnarray}
\cL_c(\bstheta;\bY,\bz,\bt)
&=& \sum_{i=1}^{n} \sum_{k=1}^K z_{ik}\log
\left[\pi_{k}(t_i;\bw)\mathcal{N}\left(\bsy_i;\bB^T_k\bt_i,\bsSigma_k\right)\right]
\label{eq: complete log-likelihood for the rhlp}
\end{eqnarray}
where $z_{ik}$ is an indicator-binary variable such that $z_{ik} = 1$ if $z_i=k$ (i.e., when $\bsy_i$ is generated by the $k$th regression model), and $0$ otherwise. 
The next section presents the dedicated EM algorithm for the multiple regression model with hidden logistic process (MRHLP).

\subsection{The dedicated EM algorithm}
\label{ssec. EM algortihm}

The proposed EM algorithm starts with an initial parameter $\bstheta^{(0)}$ and alternates between the two following steps until convergence:

\paragraph{E-Step}

This step consists of computing the expectation of the complete-data log-likelihood (\ref{eq: complete log-likelihood for the rhlp}) 
, given the observations and the current value $\bstheta^{(q)}$ of the parameter $\bstheta$ ($q$ being the current iteration):
\begin{eqnarray}
\!\!\!\!\!\!\!\! Q(\bstheta,\bstheta^{(q)})& \!\!\! = \!\!\! & \E\left[\cL_c(\bstheta;\bY,\bt,\bz)|\bY,\bt;\bstheta^{(q)}\right]\nonumber\\ 
& \!\!\! = \!\!\!   &\sum_{i=1}^{n}\sum_{k=1}^K \tau^{(q)}_{ik}\log \pi_{k}(t_i;\bw) \! + \! \sum_{i=1}^{n}\sum_{k=1}^K \tau^{(q)}_{ik}\log \N \left(\bsy_i;\bB^T_k\bt_i,\bsSigma_k \right),
\end{eqnarray}
where:
\begin{eqnarray}
 \tau^{(q)}_{ik} &=& \E [z_{ik}|\bsy_i,t_i;\bstheta^{(q)}] = p(z_{ik}=1|\bsy_i,t_i;\bstheta^{(q)}) \nonumber \\ 
 &=& \frac{\pi_{k}(t_i;\bw^{(q)})\N(\bsy_i;\bB^{T(q)}_k\bt_i,\bsSigma^{(q)}_k)}
{\sum_{\ell=1}^K\pi_{\ell}(t_i;\bw^{(q)})\N(\bsy_i;\bB^{T(q)}_{\ell}\bt_i,\bsSigma^{(q)}_{\ell})}\;
\label{eq: posterior prob tau_ik}
\end{eqnarray}
is the posterior probability that $\bsy_i$ originates from the $k$th polynomial regression model that describes the $k$th activity.
\\As shown in the expression of the $Q$-function, this step simply requires the computation of the posterior probabilities $\tau^{(q)}_{ik}$.

\paragraph{M-Step}

In this step, the value of the parameter $\bstheta$ is updated by computing the parameter $\bstheta^{(q+1)}$ maximizing the conditional expectation $Q$ with respect to $\bstheta$:
\begin{equation}
\bstheta^{(q+1)} =  \arg \max_{\bstheta} Q(\bstheta,\bstheta^{(q)}) \cdot
\end{equation}  
Let us denote by $Q_\bw(\bw,\bstheta^{(q)})$ the term in $Q(\bstheta,\bstheta^{(q)})$ that is function of $\bw$ and by $Q_{\bstheta_k}(\bstheta,\bstheta^{(q)})$ the term in $Q(\bstheta,\bstheta^{(q)})$ that depends on $\bstheta_k = (\bB_k,\bsSigma_k)$, Thus:
\begin{equation}
Q(\bstheta,\bstheta^{(q)})=Q_\bw(\bw,\bstheta^{(q)})+\sum_{k=1}^K Q_{\bstheta_k}(\bstheta,\bstheta^{(q)}),
\end{equation}
where:
$$Q_\bw(\bw,\bstheta^{(q)})=\sum_{i=1}^{n}\sum_{k=1}^K \tau^{(q)}_{ik}\log \pi_{k}(t_i;\bw)$$ 
and 
$$Q_{\bstheta_k}(\bstheta,\bstheta^{(q)}) = \sum_{i=1}^{n} \tau^{(q)}_{ik}\log \N\left(\bsy_i;\bB^T_k\bt_i,\bsSigma_k\right)$$
for $k=1,\ldots,K$. Thus, the maximization of $Q$ with respect to $\bstheta$ can be performed by
separately maximizing $Q_\bw(\bw,\bstheta^{(q)})$ with respect to $\bw$ and
$Q_{\bstheta_k}(\bstheta,\bstheta^{(q)})$ with respect to $(\bB_k,\bsSigma_k)$
for all $k=1,\ldots,K$.  
Maximizing $Q_{\bstheta_k}(\bstheta,\bstheta^{(q)})$ with respect to $\bB_k$ consists of solving the weighted least-squares problem where the weights are the posterior probabilities $\tau^{(q)}_{ik}$.  
 The solution to this problem is obtained in a closed form from the so-called normal equations and is given by: 
\begin{eqnarray}
{\bB}_k^{(q+1)} &=& (\bX^T\bW_k^{(q)}\bX)^{-1}\bX^T\bW_k^{(q)}\bY,
\label{eq: EM estimate of beta_k for the RHLP}
\end{eqnarray}
where $\bW_k^{(q)}$ is an $n \times n$ diagonal matrix of weights whose diagonal elements are the posterior probabilities $(\tau_{1k}^{(q)},\ldots,\tau_{nk}^{(q)})$ and $\bX$ is the $n\times (p+1)$ regression matrix.
Maximizing $Q_{\bstheta_k}(\bstheta,\bstheta^{(q)})$ with respect to $\bsSigma_k$ is  a weighted variant of the problem of estimating the variance of a multivariate Gaussian density. The problem can be solved in a closed form and the solution is given by:  
\begin{eqnarray}
{\bsSigma}_k^{(q+1)} &=& \frac{1}{\sum_{i=1}^{n} \tau^{(q)}_{ik}}(\bY - \bX \bB_k^{(q+1)})^T \bW_k^{(q)} (\bY - \bX \bB_k^{(q+1)}).
\label{eq: EM estimate of sigma^2_k for the RHLP}
\end{eqnarray}

The maximization of $Q_\bw(\bw,\bstheta^{(q)})$ with respect to $\bw$ is a multinomial
logistic regression problem weighted by $\tau^{(q)}_{ik}$ which we
solve with a multi-class Iterative Reweighted Least Squares (IRLS)
algorithm \citep{irls,chen99,krishnapuram, chamroukhi_et_al_NN2009}. 

The time complexity of the E-step of this EM algorithm is of $\cO(K n)$.  
The calculation of the regression coefficients in the M-step requires the computation and the inversion of the square matrix $\bX^T \bX$ which is of dimension $d\times(p+1)$, and a multiplication by the observation sequence of length $m$ which has a time complexity of $\cO(d^2(p+1)^2n)$. In addition, each IRLS loop requires an inversion of the Hessian matrix which is of dimension $2\times(K-1)$. The complexity of the IRLS loop is then approximatively of  $\cO(I_\text{IRLS} K^2)$ where $I_{\text{\tiny{IRLS}}}$ is the average number of iterations required by the internal IRLS algorithm. The proposed algorithm has therefore a time complexity of  $\cO(I_{\text{\tiny{EM}}} I_{\text{\tiny{IRLS}}} K^3 d^2p^2 n)$, where $I_{\text{\tiny{EM}}}$ is the number of
iterations of the EM algorithm. 
The pseudo code \ref{algo: proposed algorithm rhlp} summarizes one run of the proposed EM algorithm.
\begin{algorithm}[H]
\caption{\label{algo: proposed algorithm rhlp} Pseudo code of the proposed algorithm for the MRHLP model.}
{\bf Inputs:} time series $\bY$, number of polynomial components $K$, polynomial degrees $p$,  
sampling time $t_1,\ldots,t_n$.
\begin{algorithmic}[1]
 \STATE \textbf{Initialize:} $\bstheta^{(0)}= (\bw^{(0)}, \bB_1^{(0)},\ldots,\bB_K^{(0)}, \bsSigma_1^{(0)},\ldots,\bsSigma_K^{(0)})$
\STATE fix a threshold $\epsilon>0$, set $q \leftarrow 0$ (EM iteration)
 \WHILE {increment in log-likelihood $> \epsilon$}
\STATE \begin{verbatim} // E-step:
\end{verbatim} 
		\FOR{$k=1,\ldots,K$}
		\STATE compute $\tau_{ik}^{(q)}$ for $i=1,\ldots,n$ using equation (\ref{eq: posterior prob tau_ik})
	\ENDFOR
\STATE \begin{verbatim} // M-step: 
\end{verbatim} 
\FOR{$k=1,\ldots,K$}
	\STATE compute $\bB_k^{(q+1)}$ using equation (\ref{eq: EM estimate of beta_k for the RHLP})
	\STATE compute $\bsSigma_k^{(q+1)}$ using equation (\ref{eq: EM estimate of sigma^2_k for the RHLP})
\ENDFOR 
\STATE compute $\bw^{(q+1)}$ using the IRLS algorithm

 \STATE $q \leftarrow q+1$
\ENDWHILE
\STATE $\hat{\bstheta}=  \bstheta^{(q)}$ 
\end{algorithmic}
{\bf Outputs:} $\hat{\bstheta}=  (\hat{\bw},\hat{\bB}_1,\ldots,\hat{\bB}_K,\hat{\bsSigma}_1,\ldots,\hat{\bsSigma}_K)$, $\pi_{k}(t_i;\hat{\bw})$
\end{algorithm}

Once the model parameters are estimated by the EM algorithm, the time series segmentation can  be obtained by computing the estimated label $\hat{z}_i$ of the polynomial regime generating each measurement $\bsy_i$. This can be achieved by maximizing the following probability, that is:
\begin{equation}
\hat{z}_i = \arg \max_{1\leq k \leq K} \pi_{k}(t_i;\hat{\bw}), \quad  (i=1,\ldots,n).
\label{eq:  estimated segment label with the rhlp}
\end{equation}
Note that when the logistic model (\ref{eq: multinomial logit}) for $\pi_k$  is stated with one dimensional variable (i.e., if the probabilities $\pi_{k}$ are computed with a dimension $u=1$ of $\bsw_k$ $(k=1,\ldots,K)$), applying this rule guarantees that the time series are segmented into contiguous segments. 

\subsection{Selecting the number of activities} 
The proposed unsupervised approach does not require any annotation of the raw acceleration data by experts to learn the model parameters. However, in its current formulation, we assume that the number of activities is known and therefore supplied by the user (in practice this can for example be given by an expert). In a general use of the proposed model,  namely when mining amount of data for which no information regarding the activities is available, the optimal  value of $K$ can be estimated automatically from the data by using for example the Bayesian Information Criterion (BIC) \citep{BIC} which is a penalized
likelihood criterion, defined as:
\begin{equation}
\mbox{BIC}(K,p,u) = \cL(\hat{\bstheta}) - \frac{\nu_{\bstheta} \log(n)}{2}\enspace,
\end{equation}
where $\nu_{\bstheta} = \nu_{\bstheta} = K(p+4)-2$ is the number of free parameters of the model and $\cL(\hat{\bstheta})$ is the observed-data log-likelihood obtained at convergence of the EM algorithm. 

\section{Experimental study} 

In this section, the application of the proposed approach is illustrated on real time series of acceleration data measured during human activities. The  model performances are compared to those obtained with alternative activity recognition approaches. The evaluation criterion is the error segmentation (classification) between the obtained segmentation and the ground truth.  In the following subsection, we describe the experimental set up.
  
 \subsection{Experimental Setup}
\label{ssec: experimental setup}
 
\begin{figure}[H]
\centering
\includegraphics[height=4.5cm]{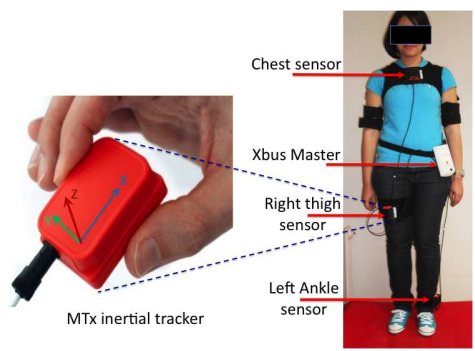}
\caption{MTx-Xbus inertial tracker and sensors placement}
\label{fig_MTX}
\end{figure}

The sensors used for data acquisition consisted of three MTx 3-DOF inertial trackers developed by Xsens Technologies \citep{Xsens}. Each MTx unit includes a tri-axial accelerometer measuring the acceleration in the 3-d space (with a dynamic range of $\pm$5$g$ where $g$ represents the gravitational constant). The sensor's placements is chosen to represent the human body motion while guaranteeing less constraint and better comfort for the wearer as well as its security. The sensors were placed at the chest, the right thigh and the left ankle respectively as shown in Figure \ref{fig_MTX}. As shown in \cite{Bouten}, points near the hip and torso exhibit a $6g$ range in acceleration. Our experiences show also that the measured ankle-sensor accelerations during the different activities do not excceed the limit of $\pm$ $5g$ . The sampling frequency is set to $25$ Hz, which is sufficient and larger than $20$ Hz the required frequency to assess daily physical activity \citep{Bouten}. The sensors were fixed on the subjects with the help of an assistant before the begining of the measurement operation. Sensors placement is chosen to represent predominantly upper-body activities such as standing up, sitting down, etc. and predominantly lower body activities such as walking, stair ascent, stair descent, etc. To secure each MTx unit in place, specific straps are used. This combination allows for efficient inter-subject transfer. The MTx units are connected to a central unit called Xbus Master which is attached to the subject's belt. Raw acceleration data are collected over time when performing the activities and the data transmission between units and the receiver is carried out through a Bluetooth wireless link.

\begin{figure}[H]
\centering
\includegraphics[height=4.5cm]{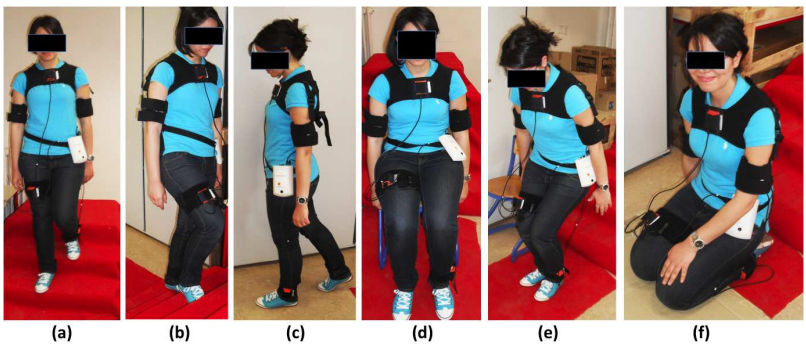}
\caption{Examples of some considered activities: a) Stair descent, b) Stair ascent, c) Walking, d) Sitting, e) Standing up, f) Sitting on the ground.}
\label{fig:Activities}
\end{figure}

The experiments were conducted at the LISSI Lab/University of Paris-Est Cr\'eteil (UPEC) by six different healthy subjects of different ages (who are not the researchers) in the office environment. In order to gather representative dataset, the recruited volunteers subjects have been chosen in a given marge of age (25-30) and weight (55-70)kg. Activity labels were estimated by an independent operator. Data are stored on a file and acceleration signals are analyzed using MATLAB software. 
Twelve activities and transitions were studied and are shown in Table \ref{table_I} and some of these activities are illustrated on Figure \ref{fig:Activities}. 
\begin{table}[H]
\centering
{\small \begin{tabular}{|c|c|}
\hline
\bfseries Activity reference & \bfseries Description\\
\hline
A$_1$ & Stair descent\\
\hline
A$_2$ & Standing\\
\hline
A$_3$ & Sitting down\\
\hline
A$_4$ & Sitting\\
\hline
A$_5$ & From sitting to sitting on the ground\\
\hline
A$_6$ & Sitting on the ground\\
\hline
A$_7$ & Lying down\\
\hline
A$_8$ & Lying\\
\hline
A$_9$ & From lying to sitting on the ground\\
\hline
A$_{10}$ & Standing up\\
\hline
A$_{11}$ & Walking\\
\hline
A$_{12}$ & Stair ascent\\
\hline
\end{tabular}}
\caption{Description of the considered activities.}
\label{table_I}
\end{table}

The activities were chosen to have an appropriate representation of everyday activities involving different parts of the body. The recognized activities and transition differ in duration and intensity level.
The subjects were asked to perform the activities in their own style and were not restricted on how the activities should be performed but only with the sequential activities order.  
Note that the activities A$_3$, A$_5$, A$_7$, A$_9$ and A$_{11}$  represent dynamic transitions between static activities.

For the three sensor units, each unit being a tri-axial accelerometer, a 9-dimensional acceleration time series  are recorded overtime for each activity. The time series present regime changes over time, in which each regime is associated to an activity. Figure \ref{fig : Acc data for illustration} illustrates an example of three acceleration data measured on the chest inertial sensor for the three activities' scenario: standing - transition (standing to sitting) - sitting - (transition) sitting to standing - standing. 
 
\subsection{Results and discussion}

 In this section, we present and discuss the results obtained by applying the EM algorithm of the proposed MRHLP model to the temporal segmentation of the acceleration time series. The data are acquired according to the experimental setup described previously. First, results are given on some activities for illustration and then the numerical results are given for the whole activities according to an experimental protocol which will be described subsequently.

\begin{figure}[H]
\centering
\begin{tabular}{cc}
\includegraphics[width=6cm]{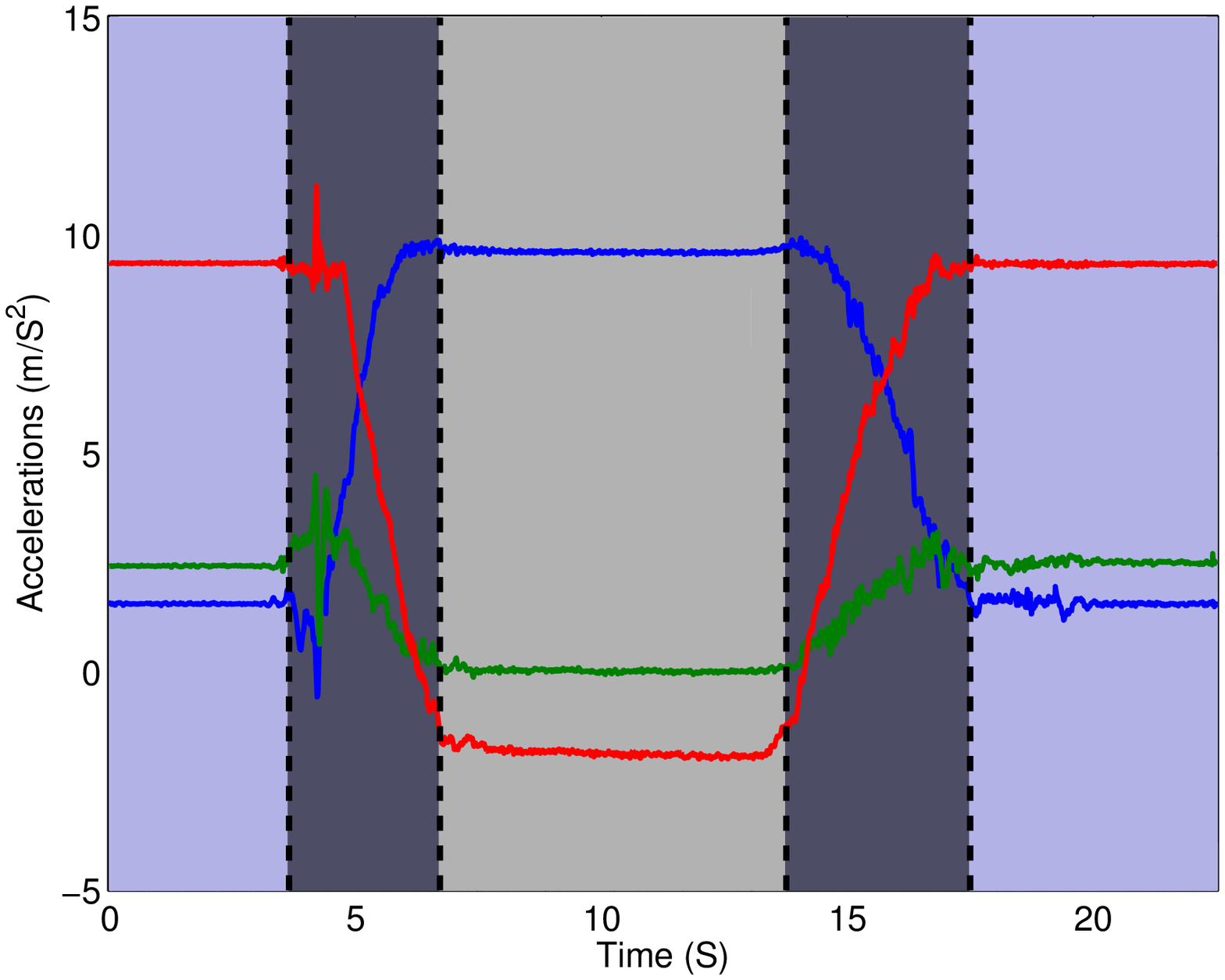}&
\includegraphics[width=6cm]{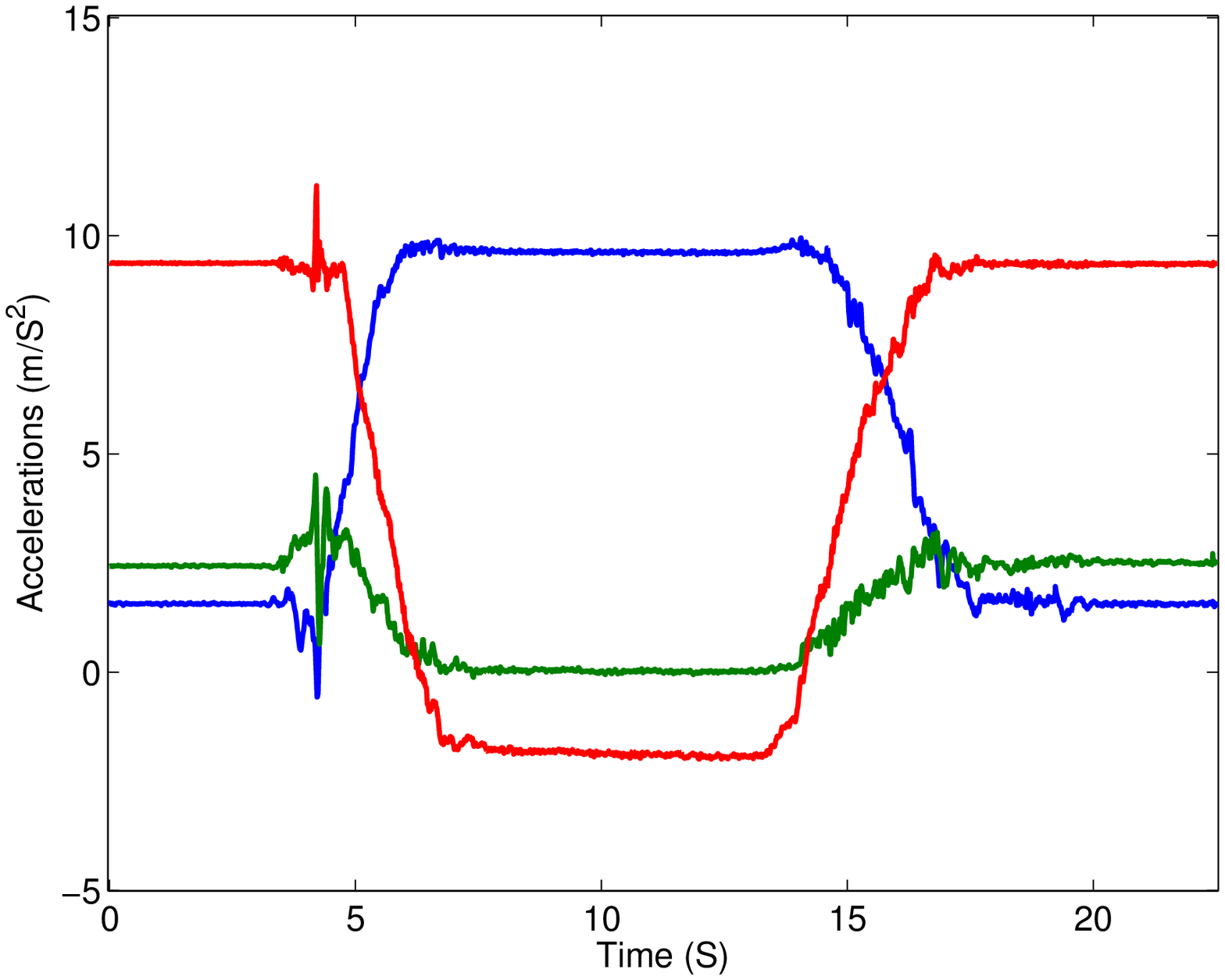}\\
\small{(a)}&\small{(b)}\\
\includegraphics[width=6cm]{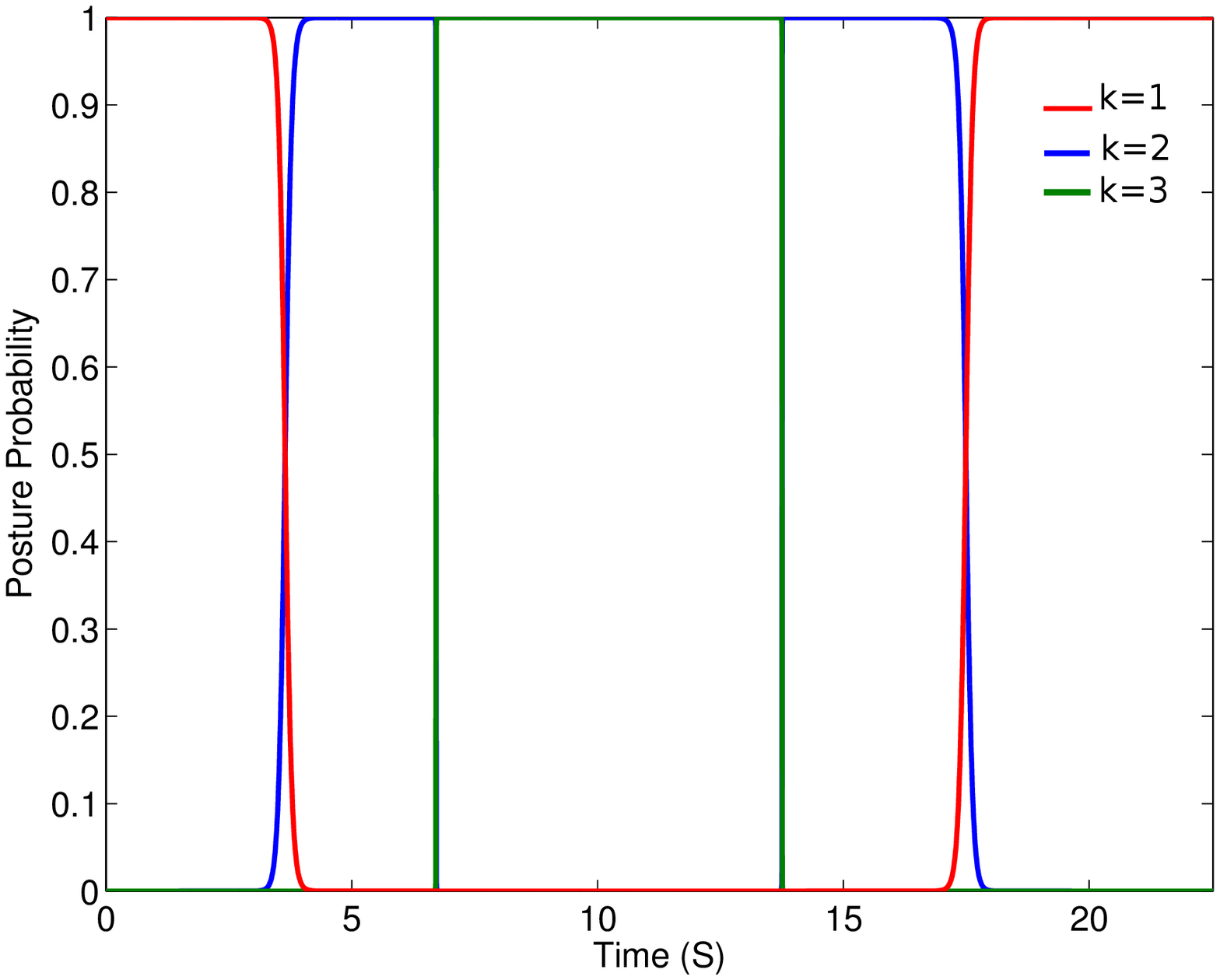} &
\includegraphics[width=6cm]{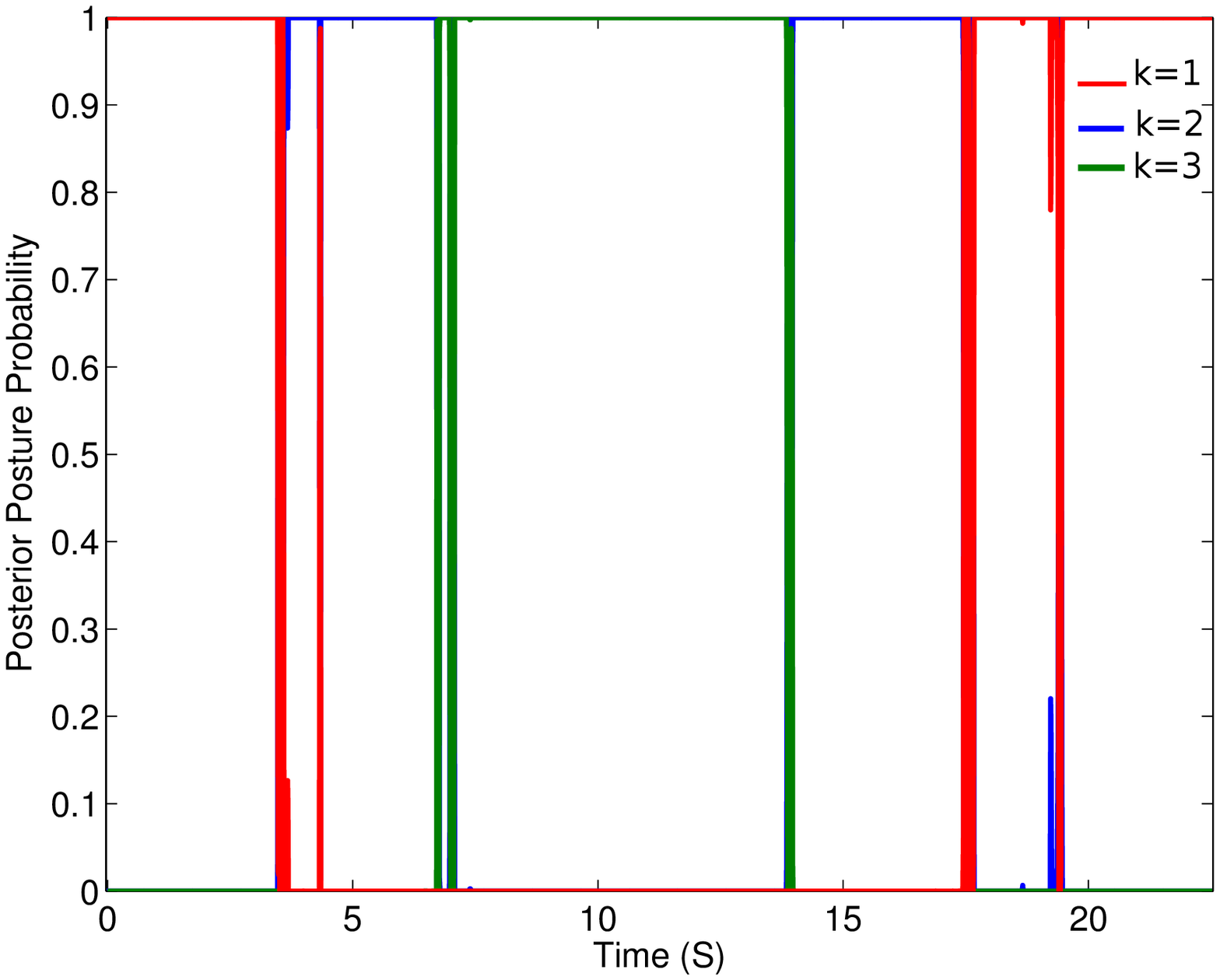}\\
\small{(c)}&\small{(d)}\\
\end{tabular}
\caption{\label{fig: illustration of MRHLP and HMM results with K=3}Results obtained by applying the proposed  MRHLP model (left) and the HMM approach (right) on the acceleration time series measured from human activity scenario shown in Figure \ref{fig : Acc data for illustration} with $k=1$: standing $k=2$: transition between standing and sitting, and $k=3$: sitting.}
\end{figure}

Consider the activities illustrated in Figure \ref{fig : Acc data for illustration} and consider for that an MRHLP model to automatically recognize the three activities. Note that in this scenario, we have two static activities with a transitory dynamic activity between sitting and standing ($K$=3). Figure \ref{fig: illustration of MRHLP and HMM results with K=3} shows the results obtained for both the MRHLP model and the HMM model. 

It can be observed on Figure \ref{fig: illustration of MRHLP and HMM results with K=3} (c) that the estimated probabilities of the logistic process that govern the switching from one activity to another over time correspond to an accurate segmentation of the acceleration times series. The acceleration data, segmented automatically, are shown in Figure \ref{fig: illustration of MRHLP and HMM results with K=3} (a). The recognized activities are represented with different colors. Moreover, the flexibility of the logistic process allows to get  smooth probabilities in particular for the transitions before 5 s and after 15 s. This can be beneficial in practice especially to decide whether or not  a decision can be made about the activity by fixing a threshold, for example 0.5.

On the other hand,  Figure \ref{fig: illustration of MRHLP and HMM results with K=3} (d) shows the posterior activity probabilities estimated by a HMM in which a standard homogeneous Markov process governs the latent activity sequence.  We clearly observe segmentation errors in the transitory phases and even when the person  maintains the same activity (see the standing activity before the instant 20 s).   

\begin{figure}[H]
\centering
\begin{tabular}{cc}
\includegraphics[width=6cm]{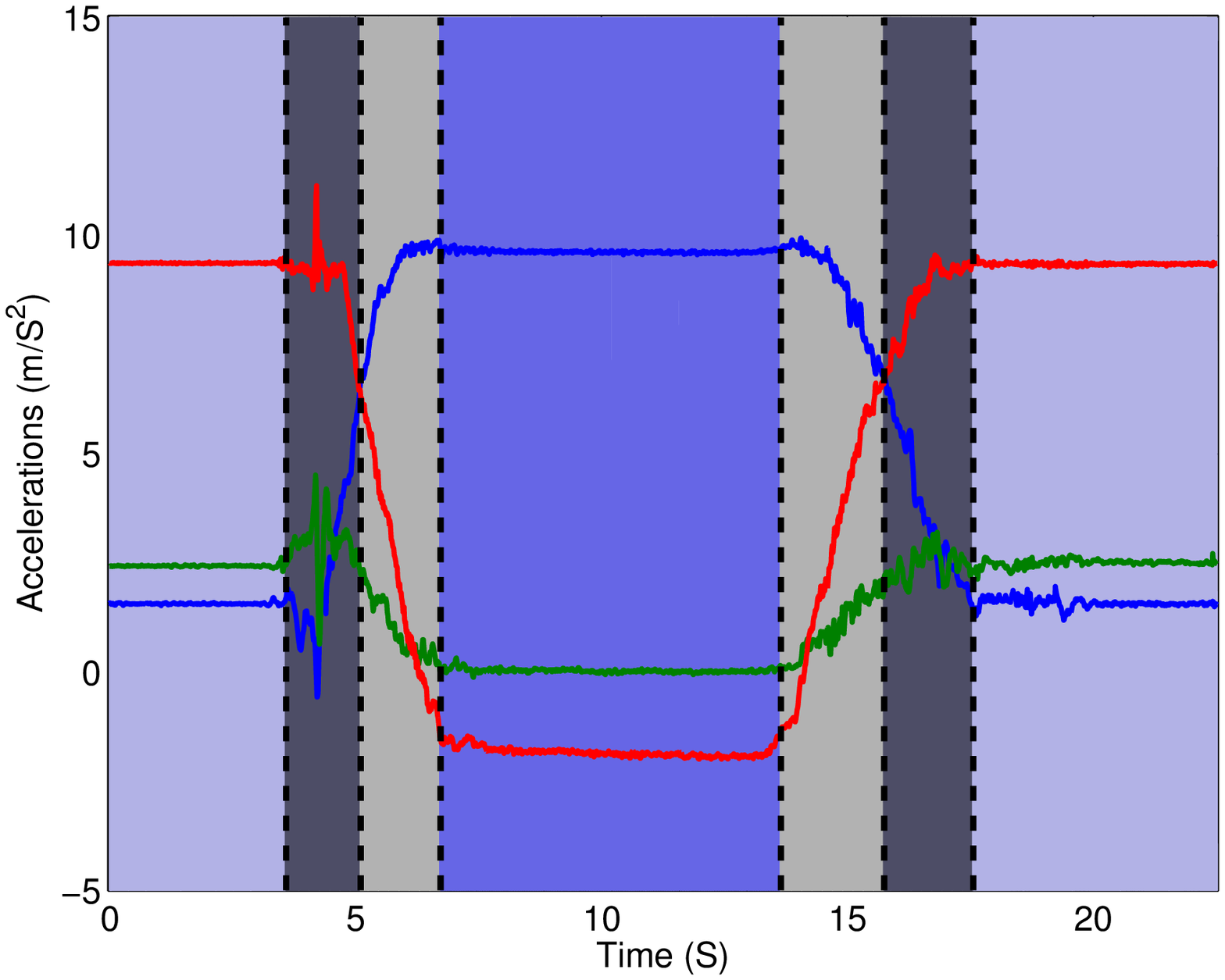}&
\includegraphics[ width=6cm]{data-set1-cuisse}\\
\small{(a)}&\small{(b)}\\
\includegraphics[width=6cm]{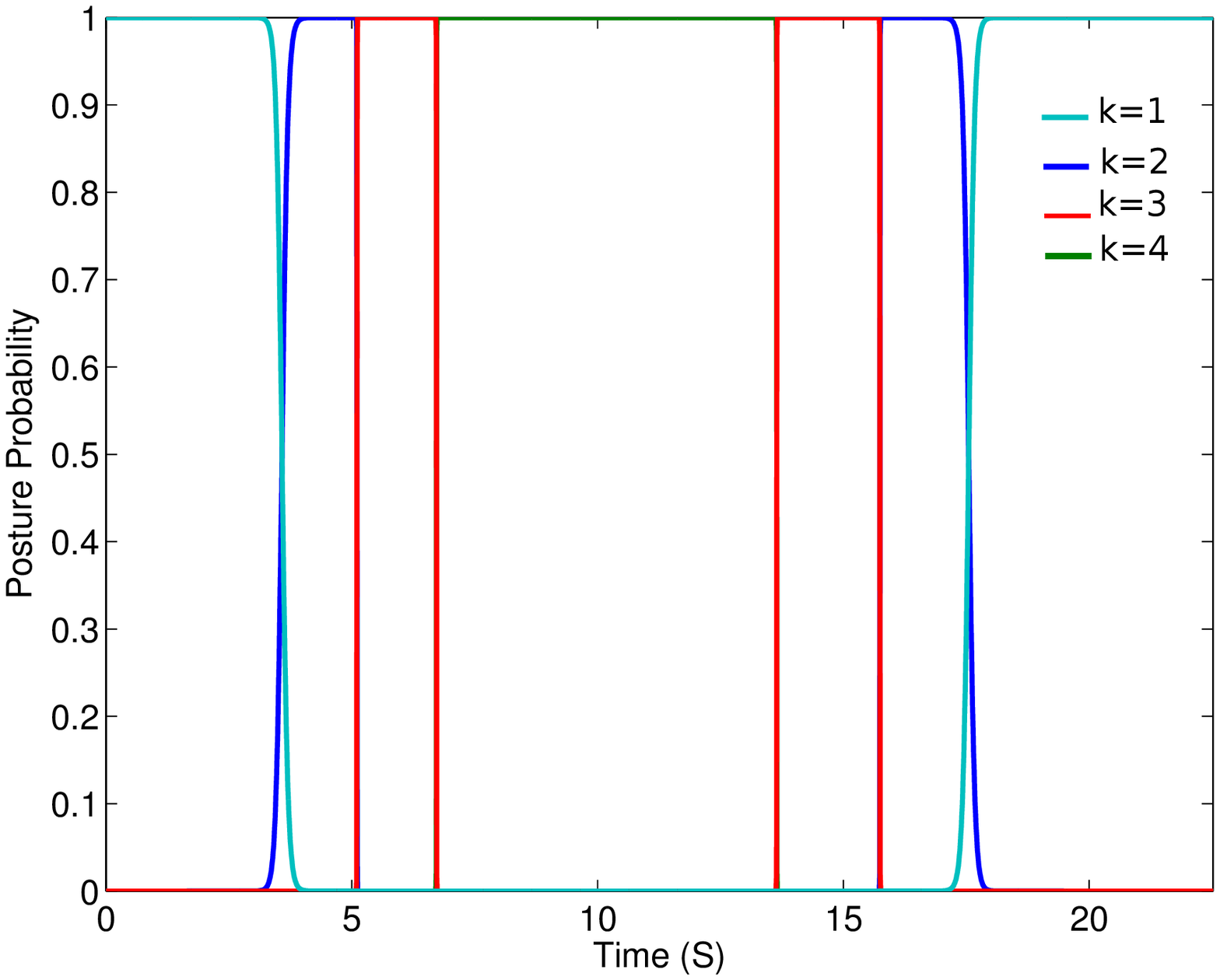} &
\includegraphics[width=6cm]{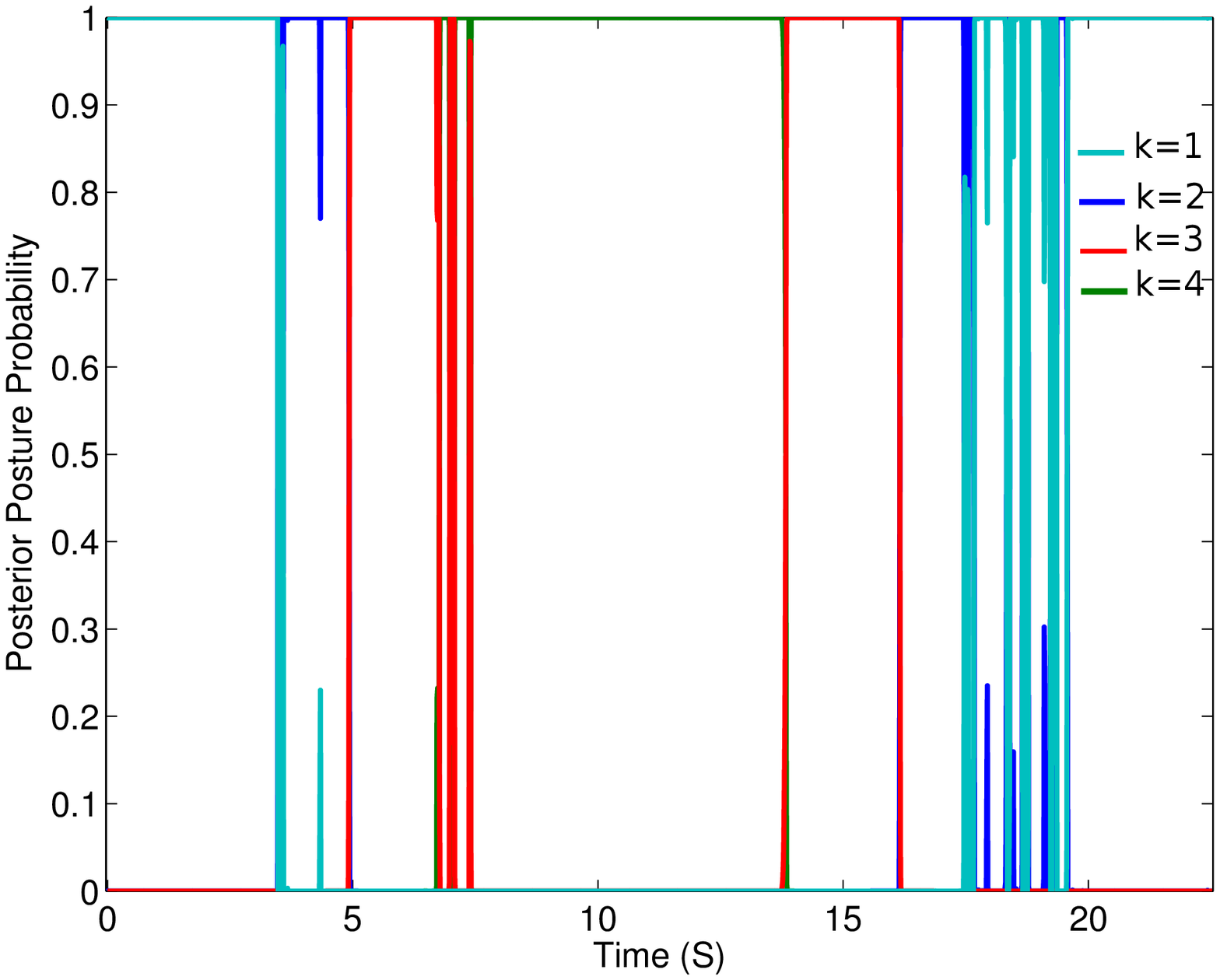}\\
\small{(c)}&\small{(d)}\\
\end{tabular}
\caption{\label{fig: illustration of MRHLP and HMM results with K=4}Results obtained by applying the proposed  MRHLP model (left) and the HMM approach (right) on the acceleration time series measured from human activity scenario shown in Figure \ref{fig : Acc data for illustration} by considering four activities with $k=1$: standing $k=2$: first pseudo transition between standing and sitting, $k=3$: second pseudo transition between standing and sitting, and $k=4$: standing.}
\end{figure}

We note that we can go more in detail in analyzing the activities by studying the transitions between activities. 
For the previous scenario, if the aim is to have more precise information on the transitions, for example to know when we are still close to the current activity or the one after, or for a more general analysis, this can be achieved by adding to the model another activity to be retrieved. This can be observed on the results presented on Figure \ref{fig: illustration of MRHLP and HMM results with K=4} and which are obtained for a scenario of four activities rather than a three activities scenario as described previously. The scenario is: standing - (transition) standing to sitting - sitting - (transition) sitting
to standing - standing. In this case, the transitory phase is analyzed as two pseudo transitions. The first transition is the one close to the previous activity and the second one is the one close to the activity to which we are going on. It can be clearly observed that the main activities still correctly segmented. Furthermore, within the transitory phases shown previously in Figure \ref{fig: illustration of MRHLP and HMM results with K=3} (a), additional segmentation is provided according to which can be seen as ``pseudo" transition within each transition (see Figure \ref{fig: illustration of MRHLP and HMM results with K=4} (a) and (c) around 5s and 15s). 
We can observe for example that the person seems to spend more time in standing up than in sitting down due to the first step of the transition to stand up which takes more time than the second step.  
The HMM approach, as shown in Figure \ref{fig: illustration of MRHLP and HMM results with K=4} provides a less satisfactory results than the proposed approach.

\begin{figure}[H]
\centering
\begin{tabular}{cc}
\includegraphics[width=6cm]{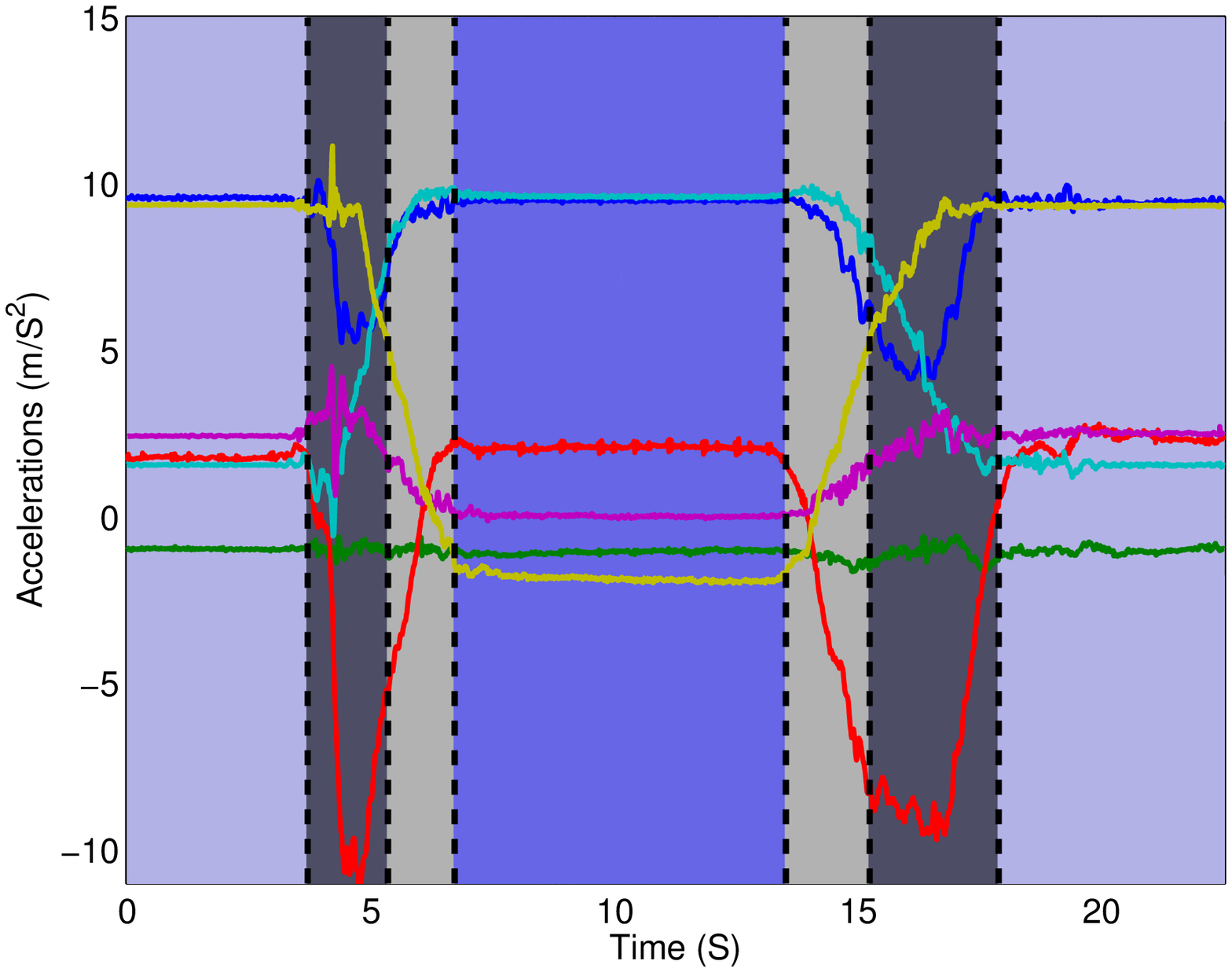}&
\includegraphics[ width=6cm]{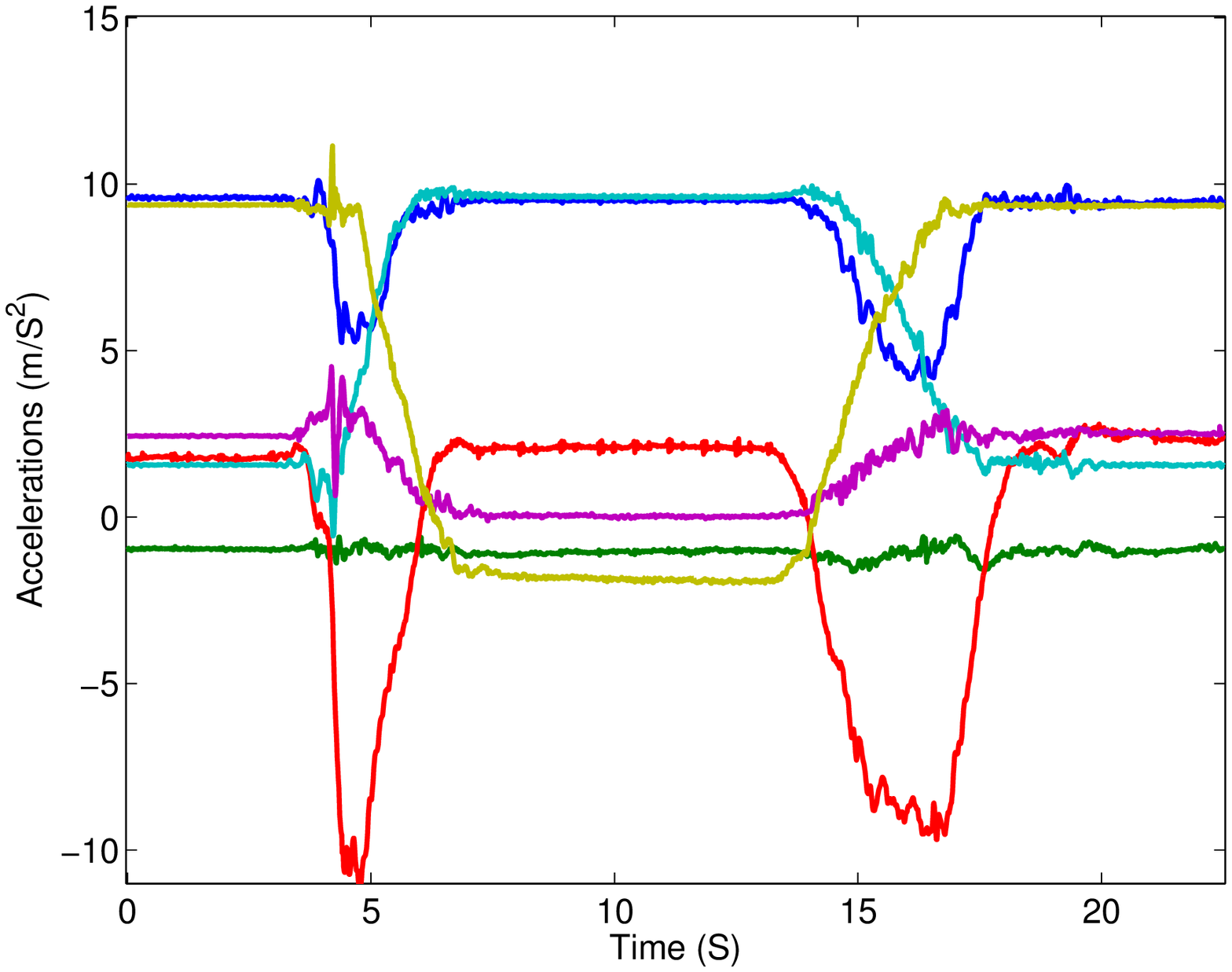}\\ 
\small{(a)}&\small{(b)}\\
\includegraphics[width=6cm]{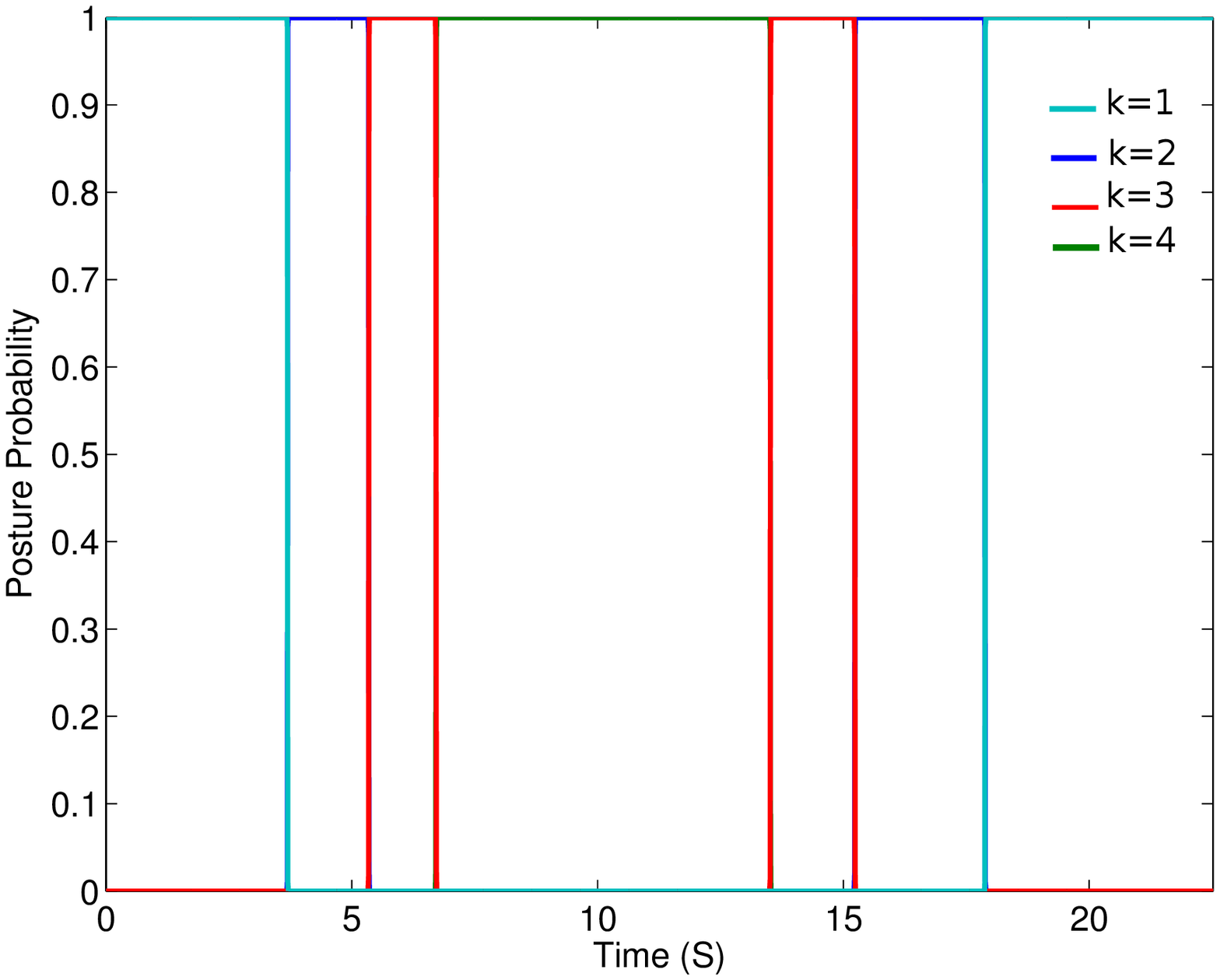} &
\includegraphics[width=6cm]{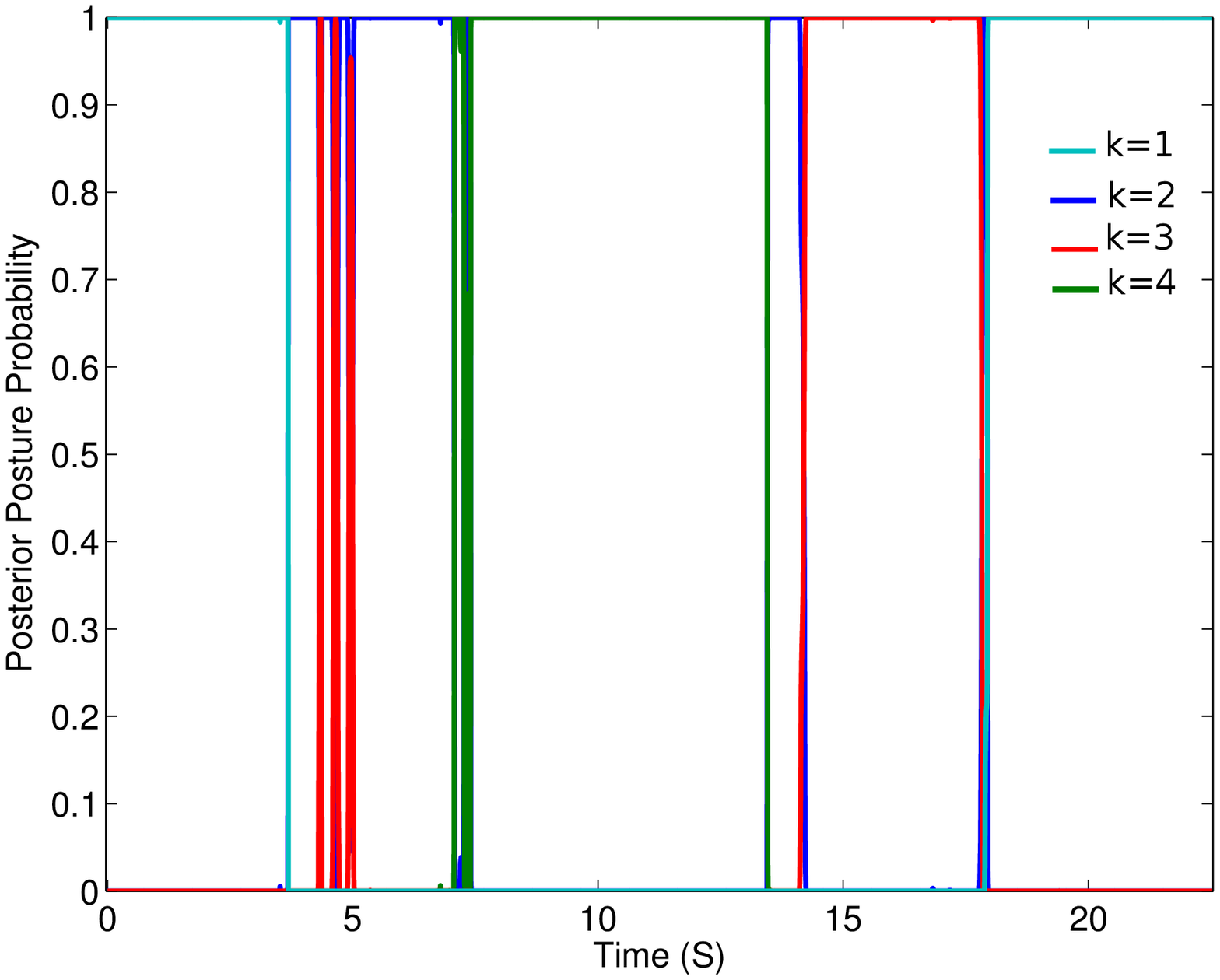}\\
\small{(c)}&\small{(d)}\\
\end{tabular}
\caption{\label{fig: illustration of MRHLP and HMM results with K=4 and d=9}Results obtained by applying the proposed  MRHLP model (left) and the HMM approach (right) on the whole 9-d acceleration time series measured from human activity scenario considered previously with $k=1$: standing $k=2$: first pseudo transition between standing and sitting, $k=3$: second pseudo transition between standing and sitting, and $k=4$: standing.}
\end{figure}

The results shown previously (see Figures \ref{fig: illustration of MRHLP and HMM results with K=3}, \ref{fig: illustration of MRHLP and HMM results with K=4}) are obtained using only
one accelerometer attached to the right thigh and providing 3-d raw acceleration times series. In Figure \ref{fig: illustration of MRHLP and HMM results with K=4 and d=9}, we illustrate the fact that, in practice, the use of the three sensors fixed at the chest, thigh and ankle, slightly improves the segmentation (both for the MRHLP and the HMM). In particular, the probabilities of the MRHLP become more close to one for the transitions, as shown in Figure \ref{fig: illustration of MRHLP and HMM results with K=4 and d=9} (c), in comparison to the previous results shown on Figures \ref{fig: illustration of MRHLP and HMM results with K=3} (c) and \ref{fig: illustration of MRHLP and HMM results with K=4} (c). Figure \ref{fig: illustration of MRHLP and HMM results with K=4 and d=9} (d) 
also shows that the segmentation is improved for the HMM compared
to the case based on a single sensor (see Figure \ref{fig: illustration of MRHLP and HMM results with K=4} (d)). Quantitative study regarding the sensors selection has been studied in \cite{Trabelsi_wIROS} and has also shown that 
using three sensors can improve the segmentation error (about 5 \%).

Now we consider the twelve activities  in an experimental protocol  to evaluate and quantify the performance of the MRHLP approach proposed in Section \ref{sec: M-rhlp model}. The automatic segmentation of the human activity is carried out by using the 3-d raw acceleration time series. The acceleration time series data, as described previously, contain recordings of six healthy volunteers performing the twelve activities described in Table \ref{table_I}. 
Each person performs the same sequence of activities.  The training was done on each data sequence for each subject. 
To judge the performance of the proposed approach, we perform comparisons to the well-known static and dynamic classification approaches such as the Naive Bayes classifier, K-NN, SVM, MLP \citep{Altun, Yang, Yang08, Preece} which are static supervised classification techniques and the HMM \citep{Jonathan-W-IROS11} which is a standard temporal segmentation approach.
We note that for the supervised static approaches and for the unsupervised dynamic HMM model, we used the same number of classes (states) as for the proposed approach, that is 12 activities; the transitions (i.e., A5 and A9) were defined as separated states. 
The dataset were divided into a training set and a test set according to a 5-fold cross validation procedure. For the supervised approaches, the classifiers were trained in a supervised way by supplying the true class labels (ground truth) in addition to the acceleration data. Then, in the test step, the class labels obtained for the test data are directly matched to the ground truth and the classification error rate is computed.
While for both the HMM approach and the proposed unsupervised MRHLP model, the models are trained in an unsupervised way from only the acceleration data, the class labels are not considered in the training, they are only used afterwards to evaluate the classifier. In the test step, as the approaches act in an unsupervised way, the class labels obtained for the test data are matched to the true labels (ground truth) by evaluating all the possible matchings; the matching providing the minimum classification error rate is selected. 

\bigskip
We notice that a main advantage of the proposed unsupervised approach is that it does not require preprocessing, the model parameters being learned in an unsupervised way from the acquired unlabelled raw acceleration data. In general, the activity recognition are two-fold as they are preceded by a preprocessing step of feature extraction from the raw acceleration data, the extracted features are then classified \citep{Altun, Ravi, Yang}. However, the feature extraction step may itself require implementing additional models or routines, well-established criteria or additional expertise to extract/select optimal features. Furthermore, the feature extraction step may also require an additional computational cost which can be penalizing in particular for a perspective of real time applications.

\begin{figure}[H]
\centering
\hspace*{.2cm}\includegraphics[width=9.75cm]{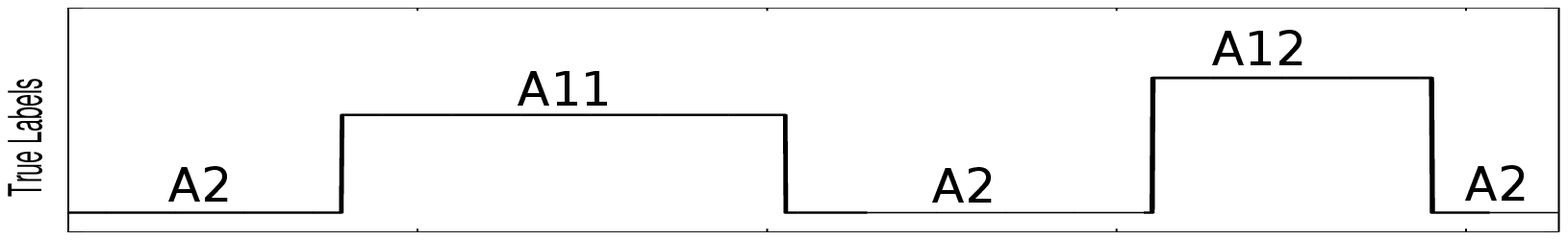}\\
\includegraphics[height = 4cm, width=10cm]{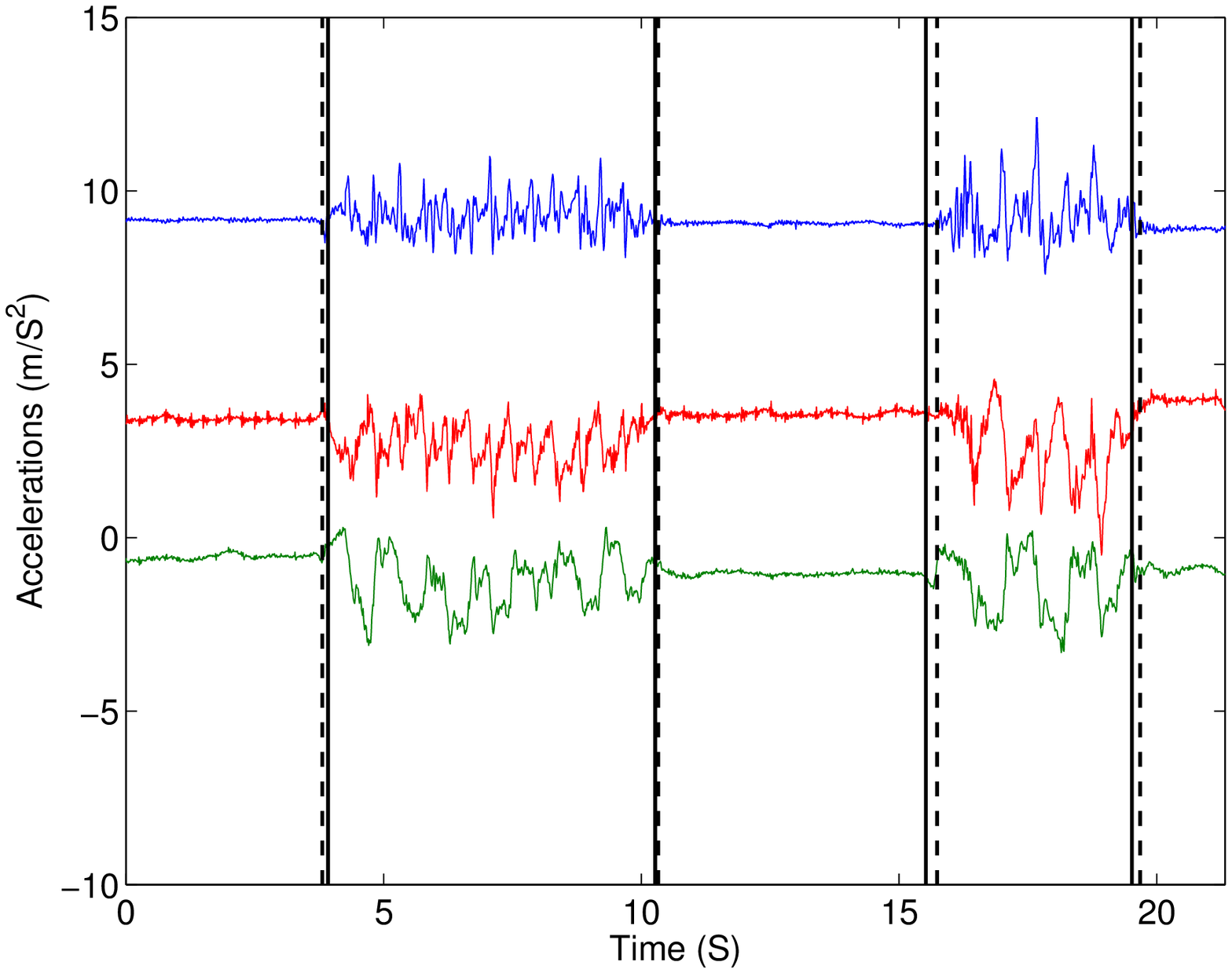}\\ 
\includegraphics[width=10cm]{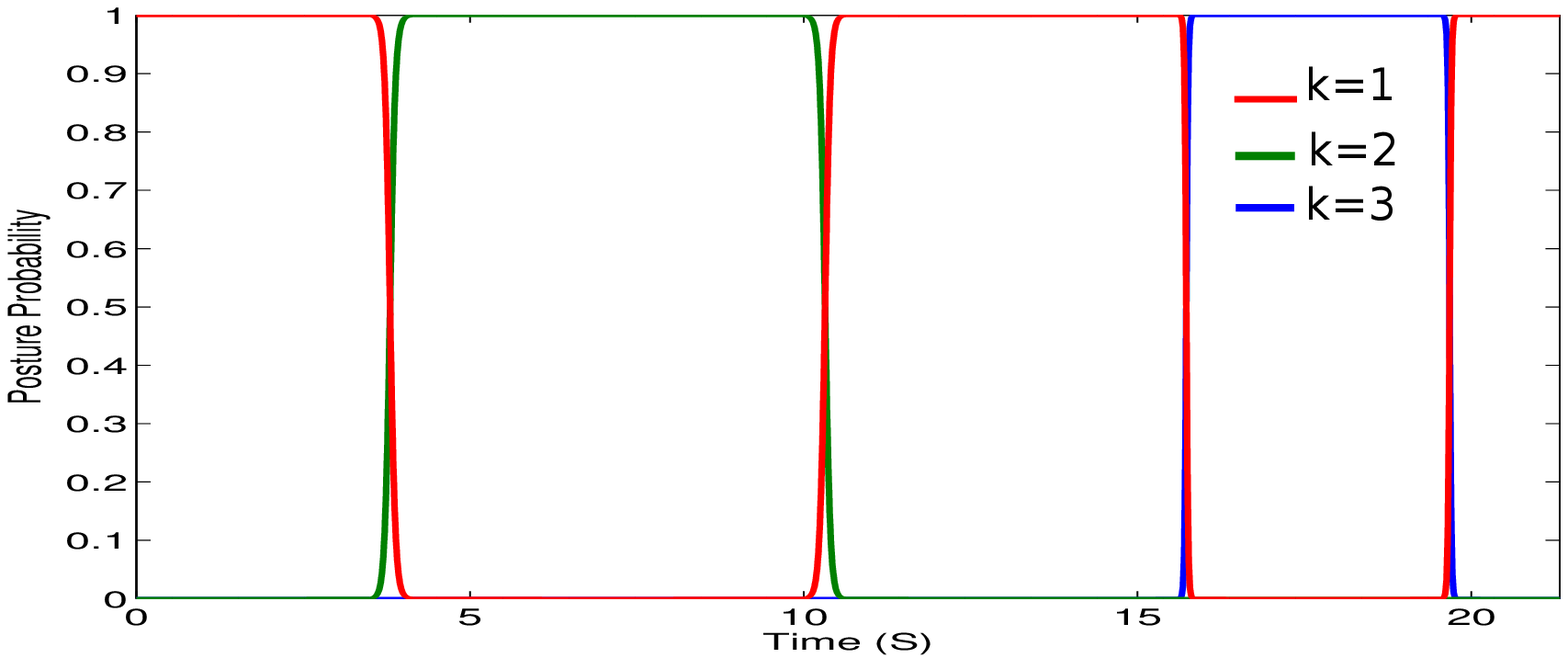}\\
\caption{ \label{fig: scenario 1} MRHLP segmentation results  for the scenario : Standing A$_2$ (k=1) - Walking A$_{11}$ (k=2) - Standing A$_2$ (k=1) - Stair ascent A$_{12}$ (k=3) - Standing A$_2$ (k=1)) with (top) the true labels, (middle) the times series and the actual segments in bold line and the estimated segments in dotted line, and (bottom) the logistic probabilities.}
\end{figure}

\begin{figure}[H]
\centering
\hspace*{.25cm}\includegraphics[height = 3cm, width=9.55cm]{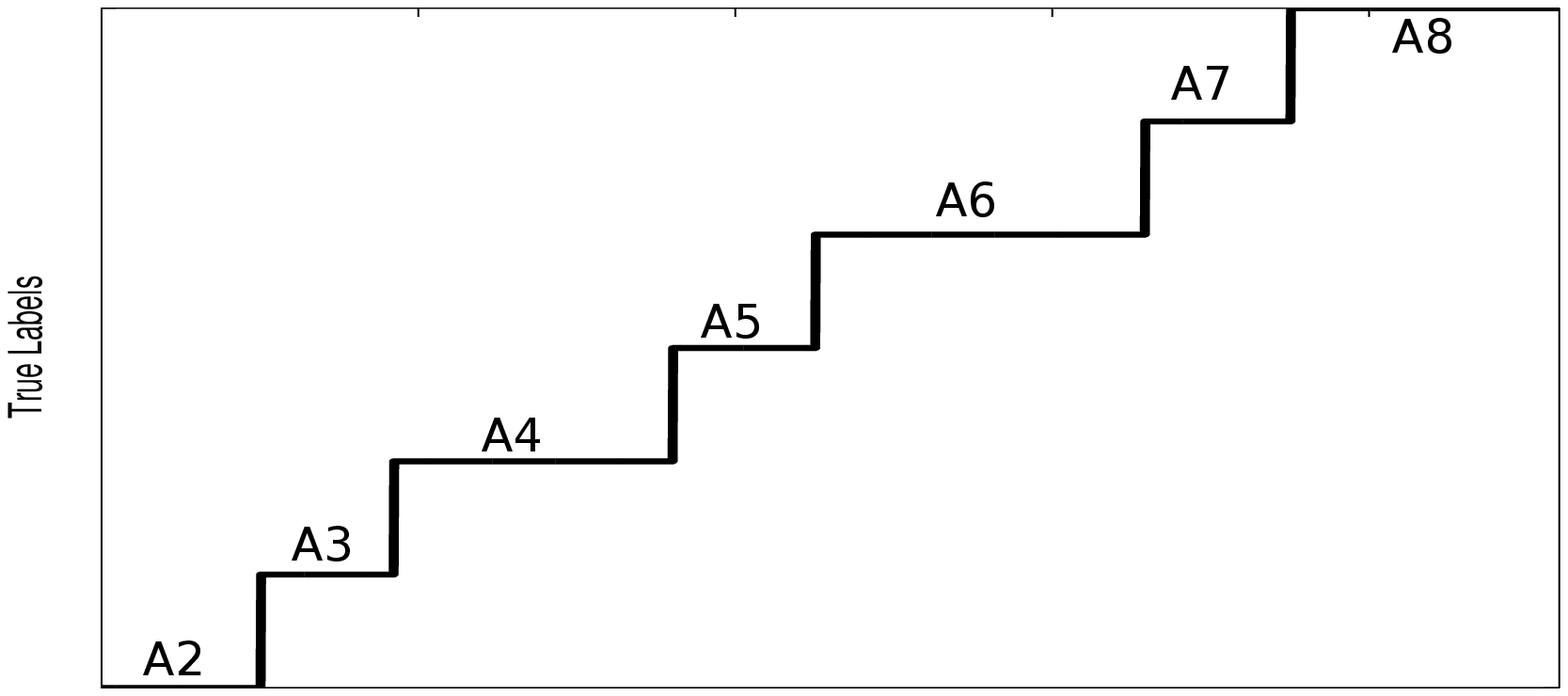}\\
\includegraphics[width=10cm]{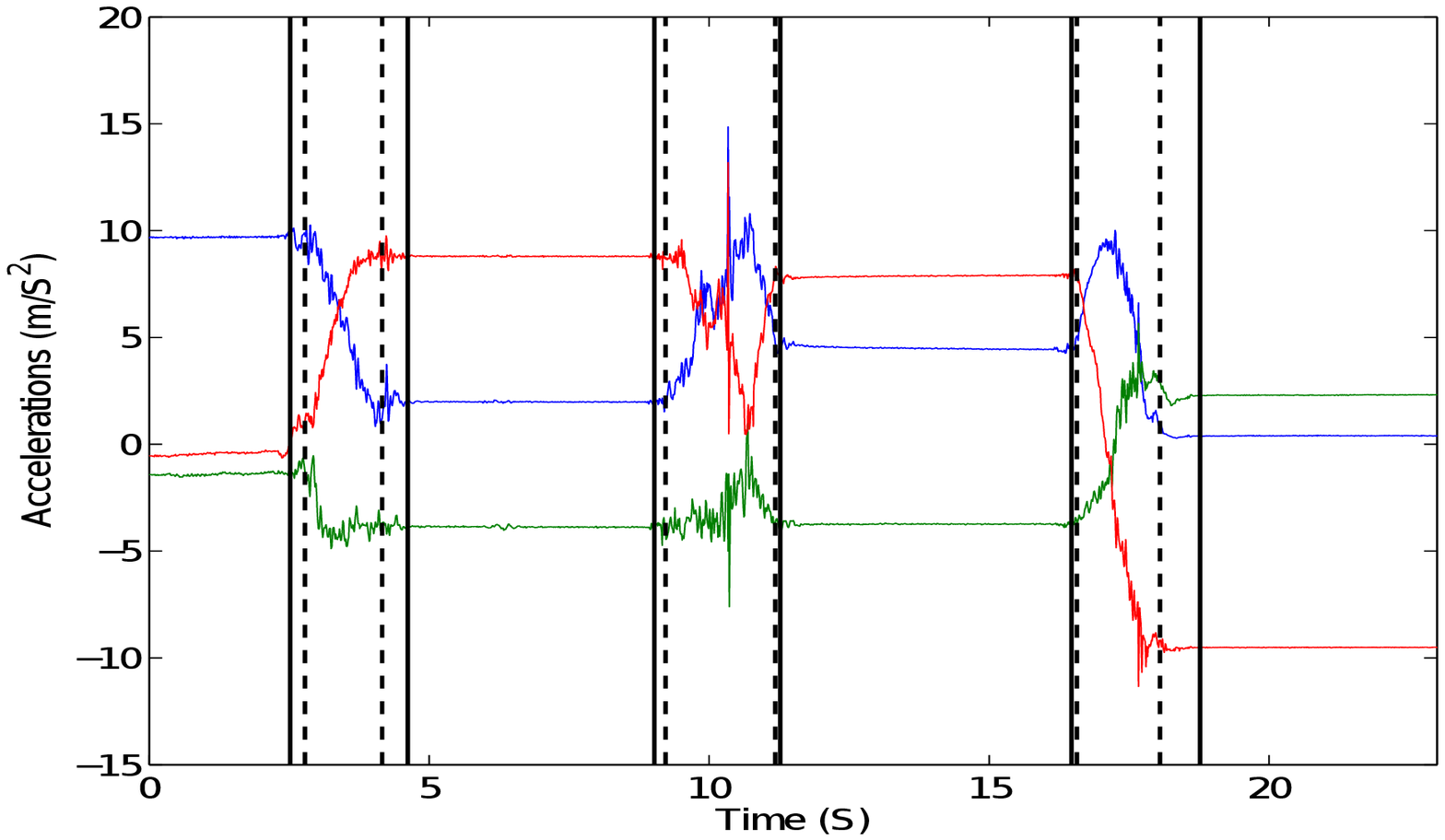}\\
\includegraphics[width=10cm]{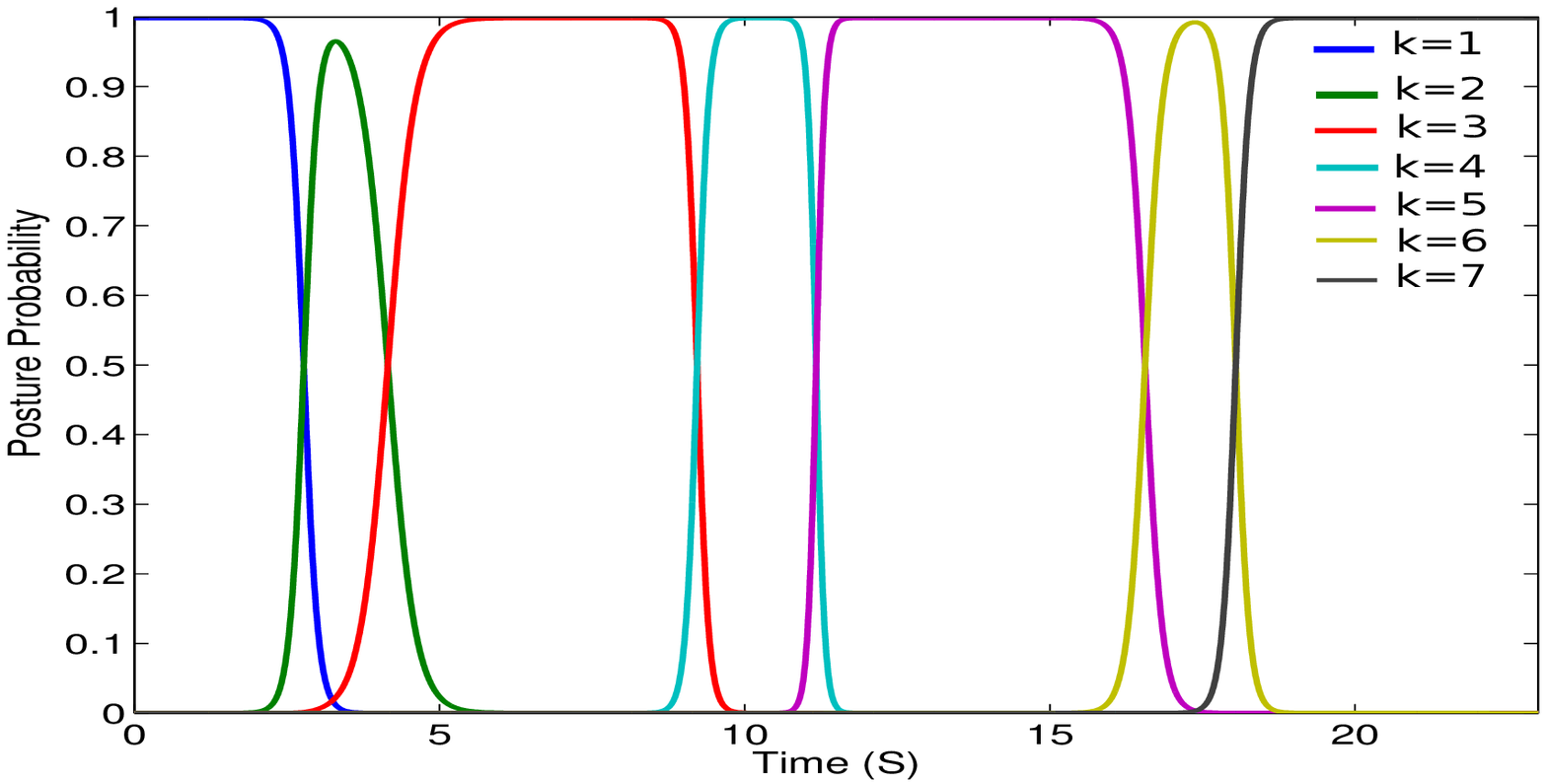}
\caption{\label{fig: scenario 2}MRHLP segmentation results for the scenario: Standing A$_2$ (k=1) - Sitting down A$_3$ (k=2)- Sitting A$_4$ (k=3) - From sitting to sitting on the ground A$_5$ (k=4) - Sitting on the ground A$_6$ (k=5) - Lying down A$_7$ (k=6) - Lying A$_8$ (k=7) with (top) the true labels, (middle) the times series and the actual segments in bold line and the estimated segments in dotted line, and (bottom) the logistic probabilities.}
\end{figure}

The results obtained with the different  approaches are given in Table \ref{tab: segmentation error rates}.   
\begin{table}[H]
\centering
\begin{tabular}{|*{1}{l||} c|}
   \hline
    Model & Correct segmentation (classification) rate ($\%$)\\
   \hline
   \hline
   Naive Bayes & $80.64$\\
   \hline
   MLP & $83.15$\\
   \hline
   SVM & $88.10$\\
   \hline
  $K$-NN & $95.89$ \\
   \hline 
    HMM & $84.16$\\
       \hline 
       \hline
    \textbf{MRHLP} & \textbf{$90.3$}\\
   \hline
\end{tabular}
\caption{\label{tab: segmentation error rates}Correct segmentation (classification) rates (\%) obtained with the different classifiers.}
\end{table}
We can observe that all the correct classification rates are greater than 80 \%. However, $K$-NN approach which provides the best results (95.8 \%) requires a significant computation time due to the computation of the distances between the considered acceleration observation and all the other observation.  
 On the other hand, the proposed approach only needs the computation of the (posterior) activity probabilities given an observation.
It can also be observed that the proposed unsupervised dynamic model provides more accurate segmentation results compared to the supervised static classifiers (Naive Bayes, MLP and SVM). This can be attributed to the fact that the model, by including the time as an intrinsic variable, fits more the temporal acceleration data. While the HMM model is also a dynamic model for time series modeling, it can be observed that it does not outperforms the proposed MRHLP approach. 
Figures \ref{fig: scenario 1}  and \ref{fig: scenario 2} show the results obtained from the proposed approach and their comparison with the ground truth segmentation.  

For the scenario of these activities, we give the confusion matrix in Table \ref{conf matrix scenario 2}.
 \begin{table}[!h]
\centering
{\small \begin{tabular}{l|l|c|c|c |c| c| c| c| c}
\multicolumn{2}{c}{}&\multicolumn{6}{c}{Obtained classes}\\
\cline{3-9}
\multicolumn{2}{c|}{}&A2&A3&A4&A5&A6&A7&A8\\
\cline{2-9}
& A2 &   245 &    7  &   0   &  0 &    0  &   0 & 0\\
\cline{2-9}
& A3 &  0 &  210  &   0  &   0  &   0  &   0  &   0\\
\cline{2-9}
True & A4 & 0  &  32 &  407  &   1  &   0  &   0  &   0\\
\cline{2-9}
classes & A5 & 0  &   0  &   0 &  225 &    0  &   0  &   0\\
\cline{2-9}
& A6 &  0  &   0  &   0  &  20 &  481  &  19   &  0\\
\cline{2-9}
& A7 & 0  &   0  &   0  &   0  &   0  & 224  &   6\\
\cline{2-9}
& A8 &  0   &  0  &   0 &    0  &   0 &    0 & 424 \\
\cline{2-9}
\end{tabular}}
\caption{\label{conf matrix scenario 2}Confusion Matrix}
\end{table}
It can be observed that the confusions of the classifier  occur in the regions of transitions between the activities. This can also be observed on the class probabilities where at these regions the probability becomes not very close to one. 

These misclassification errors can be attributed to the fact that the transitions here are similar
the problem of class overlap in the case of multidimensional data classification problem. However, the HMM approach, as it can be seen for example on the posterior classes probabilities shown in Figure \ref{fig: illustration of MRHLP and HMM results with K=3} (d), confusions can occur even within a homogeneous part of the class that is not necessarily near the transition (see the standing activity before the instant 20 s).  

The confusion matrix for the whole experiments that includes the twelve activities is given in Table \ref{conf matrix whole data}. Table \ref{FP and FN rates} gives the corresponding false positive and false negative error rates.  
We can observe from the confusion matrix that the confusions occur especially between successive activities. We can also observe that the basic activities such as A$_{12}$ and A$_4$ are more easy to detect than transitions like A$_3$.  
 \begin{table}[!h]
\hspace{-1.2 cm} {\small \begin{tabular}{l|l|c|c|c |c| c| c| c| c| c| c| c| c|}
\multicolumn{2}{c}{}&\multicolumn{10}{c}{Obtained classes}\\
\cline{3-14}
\multicolumn{2}{c|}{}& A1 & A2&A3&A4&A5&A6&A7&A8&A9&A10&A11&A12\\
\cline{2-14}
& A1 &   2356 &    122  &   13  &  0 &    0  &   0 & 0&  0 &    0  &   0 & 0& 0\\
\cline{2-14} 
& A2 &   254  &    10710 &  612 &  392 &    230  &   134 & 0&   0   &  0 &    0  &  0 & 0\\
\cline{2-14}
& A3 &   0    &    184  & 1056   &  193 &    0  &   0 & 0&   0   &  0 &    0  &   0 & 0\\
\cline{2-14}
& A4 &   0     &    20  &   92   &  2397 &  31  &   0 & 0&   0   &  0 &    0  &   0 & 0\\
\cline{2-14}

& A5 &   0 &    17  &   0   &  33 &  1486    &   30 & 0&   0   &  0 &    0  &   0 & 0\\
\cline{2-14}
Trure& A6 &   0 &    12  &   0   &  0 &    64  &   5145 & 25 &   19 & 98 & 62  &   0 & 0\\
\cline{2-14}
classes & A7 &  0 & 0  &   0  &   0  &   0  &   32  &   1688&   79   &  26 & 0  &   0 & 0\\
\cline{2-14}
& A8 & 0  &  0 &  0  &   0  &   0  &   68 &   123 &   2622   & 127 &    0  &   0 & 0\\
\cline{2-14}
& A9 & 0  &   0 &   0 &  0 &    0  &   07  &  11&   06   &  1355 &    13  &   0 & 0\\
\cline{2-14}
& A10 &  0  & 0  &   0  &  0 &  0  &  12  &  0 &   0   &  34 &    924  &   20 & 0\\
\cline{2-14} 
& A11 & 0  &  0  &   0  &   0  &   0  & 0  &   0 &   0   &  21 &    58  &   3035 & 22\\
\cline{2-14}
& A12 &  0 & 0  &   0 &    0  &   0 &    0 & 0&   0   &  0 &    64  &   421 & 2395 \\
\cline{2-14}
\end{tabular}}
\caption{\label{conf matrix whole data}Confusion Matrix for all the data set.} 
\end{table} 
\normalsize

\begin{table}[h]
\centering
{\small \begin{tabular}{|l|l|c|c|c |c| c| c| c| c| c| c| c|}
\hline 
Activity & A1 & A2&A3&A4&A5&A6&A7&A8&A9&A10&A11&A12\\
\hline 
FP rate & 9.7  & 3.2    &   40.4    & 20.4  &  17.9  &   5.2    &  8.6   & 3.8  &  18.4 &  17.5 & 12.6    & 0.9  \\
\hline
FN rate & 5.4 &  13.1    &  26.3   & 5.6  &  5.1  &   5.1   &  7.5  & 10.8  &   2.65    & 6.6   &     3.2 & 16.8   \\
\hline
\end{tabular}}
\caption{\label{FP and FN rates}False positive and false negative error rates for the MRHLP model}.
\end{table} 
\normalsize
 
We also evaluated the effect of the used sensors on the classification accuracy. We therefore performed the same experiments that involve the same activities by using different combinations of the sensors. As it can be seen on Table \ref{table: MRHLP results according to sensors}, the best results are obtained when the three sensors are used. This is also confirmed on Figure \ref{fig: illustration of MRHLP and HMM results with K=4}(c) and Figure \ref{fig: illustration of MRHLP and HMM results with K=4 and d=9}(c) that  adding sensors improves the estimation of the model since the provided activity probabilities are more precise as they are more in concordance with the true activities.
\begin{table}[!h]
\centering
\begin{tabular}{|l| c| }
   \hline
   Sensors &  Correct classification rate\\
   \hline
   \small {Chest, thigh, ankle} &   $90.3\%$ \\   
   \hline
   \small {Chest, ankle} &  $83\%$  \\
   \hline
   \small {Chest, thigh} &  $83.4\%$   \\
   \hline
   \small {Thigh, ankle} & $84.\%$    \\
   \hline
   \end{tabular}
\caption{Classification results obtained with the MRHLP model according to the considered sensors}
\label{table: MRHLP results according to sensors}
\end{table}

We note that all the approaches require a training step, except the $k$-NN which is a direct classification approach. However, the $k$-NN classifier is the more time-consuming in the test step (about 5 seconds for a single sequence). All the other approaches are not time-consuming in the test step. In the training step, the proposed algorithm and the one for the standard unsupervised HMM have a quasi-identical computing time and both are more fast than the MLP and the SVM. We also note that while the Naive Bayes is easy to train, it is outperformed by all the other approaches.    
 
In summary, the proposed approach provides a statistical well established background with very encouraging performance for automatic segmentation  of human activity. The flexibility of the model allows an accurate and efficient recognition of standard activities as well as transitory activities.

\section{Conclusion and future work}

In this paper, the problem of temporal activity recognition from acceleration data is reformulated as the one of unsupervised learning of a specific statistical latent process model for the joint segmentation of multivariate time series. The main advantages of this statistical  approach is that it  directly uses the raw acceleration data and performs in an unsupervised context which can be beneficial in practice. For example, this may be helpful for avoiding the investments in labelling flows of acceleration data or in finding adapted preprocessing feature extraction approaches. Moreover, the model formulation explicits the switching from one activity to another during time through a flexible logistic process which is also particularly well adapted for abrupt or smooth transitions. Furthermore, the expectation-maximization algorithm offers a stable efficient optimization tool to learn the model. 
The proposed MRHLP approach is applied on a real-world activity recognition problem based on multidimensional acceleration time series measured using body-worn accelerometers. The approach has shown very encouraging results compared to alternative models for activity recognition. In its current formulation, the proposed algorithm runs in a batch-mode and requires the entire data sequence to be presented. A perspective of this work is to train the proposed model with an online Expectation-Maximization (EM) algorithm (as in \cite{cappe_and_moulines_online_EM_JRSS2009}) for a real-time use perspective. 
Future work will also concern the use of other non-linear models to describe each activity signal rather than polynomial bases. This may improve in particular the representation of each activity. 
Then, another extension may consist int integrating the model into a Bayesian  non-parametric model which will be useful for any kind of complex activities and in which the number of activities will not have to be fixed.  

\bibliographystyle{elsarticle-harv}
\bibliography{references}
\end{document}